\documentclass{article}

     \usepackage[preprint, nonatbib]{nips_2018}

\usepackage[utf8]{inputenc} % allow utf-8 input
\usepackage[T1]{fontenc}    % use 8-bit T1 fonts
\usepackage{hyperref}       % hyperlinks
\usepackage{url}            % simple URL typesetting
\usepackage{booktabs}       % professional-quality tables
\usepackage{amsfonts}       % blackboard math symbols
\usepackage{nicefrac}       % compact symbols for 1/2, etc.
\usepackage{microtype}      % microtypography
\usepackage{amsfonts,amsmath,amssymb,array}
\usepackage{makeidx}         % allows index generation
\usepackage{graphicx}        % standard LaTeX graphics tool
\usepackage{multicol}        % used for the two-column index
\usepackage[bottom]{footmisc}% places footnotes at page bottom

\usepackage{times}
\usepackage{parskip}

\usepackage{wrapfig}
\usepackage{subcaption}

\usepackage{amssymb}
\usepackage{epsfig}
\usepackage{amsfonts,amsmath,amssymb,array}
\usepackage{tabularx}   % Erweiterte Tabellen Optionen
\usepackage{colortbl}
\usepackage{longtable}
\usepackage{booktabs}
\usepackage{url}
\usepackage{icomma}
\usepackage{pstricks}
\usepackage{longtable}
\usepackage{multirow}
\usepackage{color}
\usepackage{algorithm2e}

\usepackage{subfiles}
\usepackage{float}
\usepackage{mathtools}

\usepackage[sorting=none]{biblatex}
\usepackage{multirow}
\usepackage{rotating}
\usepackage{bigstrut}

\usepackage{float}

\usepackage{amsmath}

\DeclareMathOperator{\atantwo}{atan2}

\usepackage{subcaption}

\usepackage{lipsum}

\usepackage{color}

\usepackage{textcomp}

\usepackage{subcaption}

\usepackage[intoc]{nomencl}
\makenomenclature

\title{Predicting Rigid Body Dynamics using Dual Quaternion Recurrent Neural Networks with Quaternion Attention}

\author{%
  Johannes Pöppelbaum\\
  Department of Automation Technology\\
  South Westfalia University of Applied Science\\
  Soest, Germany \\
  \texttt{poeppelbaum.johannes@fh-swf.de} \\
   \And
  Andreas Schwung \\
	Department of Automation Technology\\
  South Westfalia University of Applied Science\\
  Soest, Germany \\
  \texttt{schwung.andreas@fh-swf.de} \\
}

\begin{document}

\maketitle

\begin{abstract}
We propose a novel neural network architecture based on dual quaternions which allow for a compact representation of informations with a main focus on describing rigid body movements. To cover the dynamic behavior inherent to rigid body movements, we propose recurrent architectures in the neural network. To further model the interactions between individual rigid bodies as well as external inputs efficiently, we incorporate a novel attention mechanism employing dual quaternion algebra. The introduced architecture is trainable by means of gradient based algorithms. %We apply our approach to a parcel prediction problem which exhibits complex dynamics as well as rich interactions between the parcels as well as the parcels and actuators.
We apply our approach to a parcel prediction problem where a rigid body with an initial position, orientation, velocity and angular velocity moves through a fixed simulation environment which exhibits rich interactions between the parcel and the boundaries.
\end{abstract}

\section{Introduction}

Rigid body dynamics appear in a broad range of technical applications including robotics~\cites{E.P.Lana.2015}{Pham.2018}, spacecraft dynamics~\cite{Xuan2018} and multi-body systems~\cite{Valverde.2018}. Hence, the prediction of rigid body kinematics and dynamics is particularly important for design and control of rigid bodies. Furthermore, various different applications exist in which prediction of object movements or the existence and relations of objects are relevant, like computer vision and blending \cite{LadislavKavan.2006} or neuroscience \cite{Leclercq.2013}. 

In general, different approaches for modeling rigid body transformations can be distinguished. Simple kinematics can be easily represented by employing Euler-angles. However, Euler-angles are prone to singularities. Alternatively, representation using homogeneous matrices or dual quaternions have been proposed. Their advantage lies in the combined representation of rotations and translations within a common framework. Particularly, homogeneous matrices and dual quaternions can be represented in matrix form which makes it appealing for efficient computation tools where dual quaternions have the advantage of a more compact representation of rotations and translations compared to homogeneous matrices. Consequently, different works have considered dual quaternions for robotics control and rigid body modelling and prediction.

However, in all of the mentioned approaches, the models used for prediction are fixed beforehand and hence, cannot adjust to parameter changes during runtime, unmodeled dynamics and other disturbances. Ultimately, such discrepancies between model and real world end in loss of accuracy of the prediction, especially if the prediction horizon increases. To solve these issues, models with learnable dynamics can be used. One of the most powerful group of such learnable architectures are neural networks which enjoyed great success in recent years due to the advancements in training of such networks. Particularly, recurrent neural networks are suitable for predicting general nonlinear dynamics as they appear in rigid body movements. 

As dynamic model require for some form of memory, recurrent neural networks (RNN), particularly long-short term memories (LSTM)~\cite{Hochreiter1997} and gated recurrent units (GRU) ~\cite{Cho2014} are a natural choice, but LSTM or GRU based models are general purpose function approximators without inductive bias with respect to rigid body dynamics. However, due to the specific structure of rigid body dynamics, certain form of inductive bias appears to be required for good approximations. Such forms of inductive bias are introduced within relational neural networks (RelNN)~\cite{kazemi2017relnn} or more general graph neural networks (GNN) \cites{Scarselli2009}{battaglia2018relational}{hamilton2018representation}. They rely on a graph structure, in which objects are represented by nodes while the edges model the interrelation and interactions between these different objects. 
%Both nodes and edge dynamics are modeled by neural networks. 
Different extensions are presented recently for various physical simulation objectives~\cites{li2019learning}{sanchezgonzalez2020learning}. Common to all the mentioned approaches is the introduction of strong inductive bias. Our approach is orthogonal to the mentioned approaches, as we introduce a novel form of neural network operation based on dual quaternion algebra.

In this paper, we propose a novel architecture which combines the advantages of recurrent neural network architectures and the compact representation of dual quaternion algebra to represent rigid body movements. 
Therefore, we use a hard coded numerical integration calculation where the dual quaternion encoding the position and orientation of a rigid body is incrementally updated with the help of the twist of a rigid body which causes a change in the position and orientation. This twist dual quaternion is parameterized by means of a dual quaternion neural network. Hence, the NN to calculate the transforming dual quaternion covers the rigid body dynamics and the interactions of a rigid bodie with the boundaries of the simulation environment.
%Therefore, we hard code the standard dual quaternion calculations where the dual quaternion encoding the predicted position and orientation is calculated in dual quaternion space by multiplying the dual quaternion encoding the actual position and orientation from left and right with a transformation dual quaternion and its conjugate complex. Then, this transforming dual quaternion is parameterized by means of a dual quaternion neural network which inputs the dual quaternions of other rigid bodies as well as the actuators of the considered environment. Hence, the NN to calculate the transforming dual quaternion covers the rigid body dynamics, the interactions between rigid bodies and the interaction between rigid bodies and actuators. 
As these interactions are normally sparse within an environment, we develop an attention mechanism in dual quaternion space such that the calculation of the rigid bodies dual quaternion is based only on a subset of the dual quaternion inputs. We also incorporate internal information given in scalar and vector form, particularly the rigid body parameters like length, width and depth. We employ a matrix based calculation of DQRNN which enables efficient operation due to the available efficient matrix calculation in NN. Note that the proposed architecture is agnostic to the RNN architecture, i.e. simple RNN or gated units like LSTMs and GRUs, as well as to the observation space, i.e. observation obtained from sensor fields, camera systems or internal position and velocity sensors can be integrated in the approach.

We remark that single quaternion neural networks as feedforward~\cite{T.Parcollet.2016}, including convolutional single quaternion NN~\cites{Gaudet.2017}{Zhu.2019} as well as recurrent single quaternion NN~\cite{Parcollet.2018} are well developed. However, the scopes of these papers are completely different in that these architectures are focused on an application in speech recognition or image classification/segemntation. More generally, the idea of these approaches is a compact representation of various channels of the same signals (e.g. their higher order derivatives or the rgb values of an image) or various interrelated signals which are casted into a quaternion representation. Contrary, this work focuses on rigid body dynamics which require the usage of dual quaternions to fully describe their movements which requires the development of dual quaternion NN. To the best of our knowledge this is the first approach to cast dual quaternions into end-to-end trainable NN architectures.

%We apply the proposed dual quaternion NN to a parcel transportation problem in which parcels move down an actuated inclined plane. Additional actuators are mounted on the plane to direct the parcel movements. The plane is modeled in a DEM simulation environment which provides a detailed physical model of the parcel movements and provides the training data sets. 

\section{Dual Quaternion Algebra}

In the following, initially the concept of quaternions and dual numbers is introduced, subsequently their combination to dual quaternions follows. The last paragraph addresses the usage of dual quaternions for rigid transformation. 

\subsection{Quaternions}

Quaternions were created by Hamilton in 1843 to expand the complex numbers to the three dimensional space. They can be seen as an extension to the complex numbers $\mathbb{C}$ where the complex scalar is replaced by a vector out of three scalars and three complex elements, denoted by $(i, j, k)$. Together they form a four component number
\begin{equation}
	Q = (q_{0}, \textbf{q}) = q_{0} + q_{1}i + q_{2}j + q_{3}k
\end{equation}
where $q_{0}\in\mathbb{R}$ can be referred to as the real part, whereas $\textbf{q} = (q_{1}, q_{2}, q_{3})\in\mathbb{R}$ are the imaginary elements. This leads to the set of quaternions $\mathbb{H}=\{Q : Q = q_{0} + q_{1}i + q_{2}j + q_{3}k ,~ q_{0}, q_{1}, q_{2}, q_{3}~\in~\mathbb{R}\} $. \cite{Valverde.2018}

Quaternions with a real part $q_{0}=0$ are called pure quaternions \cite{JiafengXu.2016} or vector quaternions \cite{N.Filipe.2013}. The imaginary elements $(i, j, k)$ have the properties 
\begin{equation}
\begin{split}
	i^{2} = j^{2} &= k^{2} = ijk = -1 \\
	i &= jk = -kj \\
	j &= ki = -ik \\
	k &= ij = -ji \\
\end{split}
\end{equation}
where the latter are easily derived from the first line. The conjugate $Q^{*}$ of a quaternion $Q$ is 
\begin{equation}
	Q^{*} = (q_{0}, -\textbf{q}) = q_{0} - q_{1}i - q_{2}j - q_{3}k .
\label{equ:QuatConj}
\end{equation}
Addition of two quaternions $P$ and $Q$ is defined component-wise as
\begin{equation}
	P + Q =  p_{0} + q_{0} + (p_{1} + q_{1})i + (p_{2} + q_{2})j + (p_{3} + q_{3})k
\end{equation}
and multiplication of $P$ and $Q$ as
\begin{equation}
PQ =  p_{0}q_{0} - \textbf{p}\cdot\textbf{q} + p_{0}\textbf{q} + q_{0}\textbf{p} + \textbf{p}\times\textbf{q}
\end{equation}
with the alternative representation
\begin{equation}
\begin{split}
	PQ &=  (p_{0} + p_{1}i + p_{2}j + p_{3}j)(q_{0} + q_{1}i + q_{2}j + q_{3}k) \\
	   &= (p_{0}q_{0} + p_{1}q_{1} + p_{2}q_{2} + p_{3}q_{3}) \\
	   &+i(p_{1}q_{0} + p_{0}q_{1} - p_{3}q_{2} + p_{2}q_{3}) \\
	   &+j(p_{2}q_{0} + p_{3}q_{1} + p_{0}q_{2} - p_{1}q_{3}) \\
	   &+k(p_{3}q_{0} - p_{2}q_{1} + p_{1}q_{2} + p_{0}q_{3}) .\\
\end{split}
\label{equ:quatMulLongForm}
\end{equation}
The norm of a quaternion can be calculated with 
\begin{equation}
	\left\lVert Q\right\rVert = \sqrt{(q_{0}^{2} + q_{1}^{2} + q_{2}^{2} + q_{3}^{2})} = \sqrt{QQ^{*}} = \sqrt{Q^{*}Q}  = \sqrt{Q\cdot Q}
\end{equation}
\cite{Valverde.2018}. This can be verified using equation \eqref{equ:quatMulLongForm}, $Q = q_{0} + q_{1}i + q_{2}j + q_{3}k$ and $Q^{*} = q_{0} - q_{1}i - q_{2}j - q_{3}k$:
\begin{equation}
	\begin{split}
		QQ^{*} &=  (q_{0} + q_{1}i + q_{2}j + q_{3}k)(q_{0} - q_{1}i - q_{2}j - q_{3}k) \\
		&= (q_{0}q_{0} + q_{1}q_{1} + q_{2}q_{2} + q_{3}q_{3}) \\
		&+i(q_{1}q_{0} + q_{0}(-q_{1}) - q_{3}(-q_{2}) + q_{2}(-q_{3})) \\
		&+j(q_{2}q_{0} + q_{3}(-q_{1}) + q_{0}(-q_{2}) - q_{1}(-q_{3})) \\
		&+k(q_{3}q_{0} - q_{2}(-q_{1}) + q_{1}(-q_{2}) + q_{0}(-q_{3})) \\
		&=q_{0}^{2} + q_{1}^{2} + q_{2}^{2} + q_{3}^{2} .
	\end{split}
\end{equation}
Quaternions which fulfil $\left\lVert Q\right\rVert = 1$ are called unit quaternions and form the set $\mathbb{H}_{1}=\{Q: \left\lVert Q\right\rVert = 1,~Q \in \mathbb{H} \} $. These unit quaternions can be used for rotations with a rotation angle $\theta$ and a corresponding unit rotation axis \textbf{n}, obtaining the rotation unit quaternion $Q$
\begin{equation}
	Q = (\cos\dfrac{\theta}{2}, \textbf{n}\sin\dfrac{\theta}{2}).
\label{equ:rotQuat}
\end{equation}
This quaternion can be used to rotate a point $P$ as the pure quaternion $P = p_{1}i + p_{2}j + p_{3}k$ with 
\begin{equation}
	P' = QPQ^{*}
\label{equ:RotatingWithSingleQuaternions}
\end{equation}
to the new position $P'$. \cite{BenKenwright.2013}

Alternatively to the previously showed multiplication, a convenient matrix vector multiplication can be used. For $R=PQ$ this is \cite{Valverde.2018}
\begin{equation}
	\left[\begin{array}{cccc}
		r_{1}  \\
		r_{2}  \\
		r_{3}  \\
		r_{4}  \\
	\end{array}\right]
	=
		[[P]]_{L} Q
	=
	\left[\begin{array}{cccc}
		p_{0} & -p_{1} & -p_{2} & -p_{3} \\
		p_{1} & p_{0} & -p_{3} & p_{2}  \\
		p_{2} & p_{3} & p_{0} & -p_{1} \\	
		p_{3} & -p_{2} & p_{1} & p_{0} \\
	\end{array}\right]
	\left[\begin{array}{cccc}
		q_{0}  \\
		q_{1}  \\
		q_{2} \\	
		q_{3} \\
	\end{array}\right]
\label{equ:QuatLMatrix}	
\end{equation} 
\begin{equation}
	\left[\begin{array}{cccc}
		r_{1}  \\
		r_{2}  \\
		r_{3}  \\
		r_{4}  \\
	\end{array}\right]
	=
		[[Q]]_{R} P
	=
	\left[\begin{array}{cccc}
		q_{0} & -q_{1} & -q_{2} & -q_{3} \\
		q_{1} & q_{0} & q_{3} & -q_{2}  \\
		q_{2} & -q_{3} & q_{0} & q_{1} \\	
		q_{3} & q_{2} & -q_{1} & q_{0} \\
	\end{array}\right]
	\left[\begin{array}{cccc}
		p_{0}  \\
		p_{1}  \\
		p_{2} \\	
		p_{3} \\
	\end{array}\right] .
\label{equ:QuatRMatrix}
\end{equation} 
Also the conjugates can be expressed in matrix notation:
\begin{equation}
	[[Q]]_{L}^{*}
	=
	\left[\begin{array}{cccc}
		q_{0} & q_{1} & q_{2} & q_{3} \\
		-q_{1} & q_{0} & q_{3} & -q_{2}  \\
		-q_{2} & -q_{3} & q_{0} & q_{1} \\	
		-q_{3} & q_{2} & -q_{1} & q_{0} \\
	\end{array}\right], 
	[[Q]]_{R}^{*}
	=
	\left[\begin{array}{cccc}
		q_{0} & q_{1} & q_{2} & q_{3} \\
		-q_{1} & q_{0} & -q_{3} & q_{2}  \\
		-q_{2} & q_{3} & q_{0} & -q_{1} \\	
		-q_{3} & -q_{2} & q_{1} & q_{0} \\
	\end{array}\right] .
\label{equ:QuatConjMatrix}
\end{equation}
Additionally, an exponential denoted with $exp(\cdot)$ for a quaternion exists. It can be calculated with \cite{NeilDantam.2014}% \cite{Manchester.2016}
%\begin{equation}
% 	exp(Q) 
% 	= e^{q_{0}} 
% 	\left[\begin{array}{cccc}
%	 	\cos(\left\| Q \right\| )  \\
%	 	\frac{\textbf{q}}{\left\| Q \right\|}\sin(\left\| Q \right\| )  \\
% 	\end{array}\right]
% 	= e^{q_{0}} 
% 	\left[\begin{array}{cccc}
%	 	\cos(\left\| Q \right\| )  \\
%	 	\frac{q_{1}}{\left\| Q \right\|}\sin(\left\| Q \right\| )  \\
%	 	\frac{q_{2}}{\left\| Q \right\|}\sin(\left\| Q \right\| )  \\	
%	 	\frac{q_{3}}{\left\| Q \right\|}\sin(\left\| Q \right\| )  \\
% 	\end{array}\right].
%\label{equ:quatExp}
%\end{equation}
%%
%
\begin{equation}
	exp(Q) = e^{q_{0}} \left(
		\cos(\left\| \textbf{q} \right\|) + 
		\frac{\sin(\left\| \textbf{q} \right\|)}{\left\| \textbf{q} \right\|} \textbf{q}
	\right) .
\end{equation}
Since this becomes problematic from a computational point of view when $\left\| \textbf{q} \right\| \rightarrow 0$, the following Taylor series can be used as a replacement:
\begin{equation}
	\frac{\sin(\phi)}{\phi} 
	= 1 
	- \frac{\phi^{2}}{6}  
	+ \frac{\phi^{4}}{129}
	- \frac{\phi^{6}}{5040}
	+ \dots~.
	\label{equ:singularityQuatExpTailorSeries}
\end{equation}
If the exponential of a pure quaternion is calculated, the result is always a unit quaternion \cite{JiafengXu.2016}.

As the inverse operation to the quaternion exponential, also a logarithm can be calculated in quaternion space:
%%
%\begin{equation}
%	ln(Q) 
%	=
%	\left[\begin{array}{cccc}
%		\ln(\left\| Q \right\| )  \\
%		\frac{\phi}{\left\| \textbf{q} \right\|}\textbf{q} \\
%	\end{array}\right]
%\end{equation} 
%%
%
\begin{equation}
	\ln(Q) 
	=
	\ln(\left\| Q \right\| ) + \frac{\phi}{\left\| \textbf{q} \right\|}\textbf{q} 
\end{equation} 
where
\begin{equation}
 	\phi = \cos^{-1}\left(\frac{q_{0}}{\left\| Q \right\|}\right)
 		 = \sin^{-1}\left(\frac{\left\| \textbf{q} \right\|}{\left\| Q \right\|}\right)
 		 = \atantwo\left(\left\| \textbf{q} \right\|, q_{0}\right) .
\end{equation}
When $\left\| \textbf{q} \right\| \rightarrow 0$, another Taylor series has to be used. For this, $\frac{\phi}{\left\| \textbf{q} \right\|}$ can be rewritten as
\begin{equation}
	\frac{\phi}{\left\| \textbf{q} \right\|} 
	= \dfrac{\frac{\phi}{\left\| Q \right\|}}{\frac{\left\| \textbf{q} \right\|}{\left\| Q \right\|}} 
	= \dfrac{\frac{\phi}{\left\| Q \right\|}}{\sin(\phi)}
	= \dfrac{\frac{\phi}{\sin(\phi)}}{\left\| Q \right\|}
\label{equ:singularityDuatLn}
\end{equation}
and $\frac{\phi}{\sin(\phi)}$ is approximated as a Taylor series:
\begin{equation}
	\frac{\phi}{\sin(\phi)} 
	= 1 
	+ \frac{\phi^{2}}{6}  
	+ \frac{7\phi^{4}}{360}
	+ \frac{31\phi^{6}}{15120}
	+ \dots~.
\label{equ:singularityDuatLnTailorSeries}
\end{equation}
In case of $Q$ being an unit quaternion, these calculations simplify to \cites{NeilTDantam.2018}{NeilDantam.2014}
%
%\begin{equation}
%	\ln(Q) =
%	\left[\begin{array}{c}
%		0  \\
%		\frac{\phi}{\sin(\phi)}\textbf{q} \\
%	\end{array}\right] .
%\end{equation}
\begin{equation}
	\ln(Q) = \frac{\phi}{\sin(\phi)}\textbf{q} .
\end{equation}

\subsection{Dual Numbers}

Dual numbers are comparable to complex numbers as a combination of two numbers $a, b\in\mathbb{R} $ and the dual unit $\epsilon$ which satisfies $\epsilon^{2}=0$ and $\epsilon\neq0$ such that a dual number $d$ results
\begin{equation}
	d = a + \epsilon b
\end{equation}
where $a$ is the real part and $b$ the dual part. Addition and multiplication of dual numbers is defined like in the following  \cite{Wu.2005}:
\begin{equation}
	d_{1} + d_{2} = (a_{1} + a_{2}) + \epsilon (b_{1} + b_{2} )
\end{equation}
\begin{equation}
	\begin{split}
		d_{1}d_{2} &= (a_{1}a_{2}) + \epsilon (a_{1}b_{2} + a_{2}b_{1} ) + \epsilon^{2}a_{2}b_{2}  \\
		&= (a_{1}a_{2}) + \epsilon (a_{1}b_{2} + a_{2}b_{1} ).
	\end{split}
\end{equation}
The conjugate $d^{*}$ of a dual number $d$ is \cite{Leclercq.2013}
\begin{equation}
	d^{*} = a - \epsilon b .
\end{equation}

\subsection{Dual Quaternions}

Dual quaternions combine the ideas of quaternions and dual numbers, forming a dual quaternion $Q_{d}$ out of a real part $Q \in \mathbb{H}$ and a dual part $Q_{\epsilon} \in \mathbb{H}$
\begin{equation}
	Q_{d} = Q + \epsilon Q_{\epsilon}
\end{equation}
leading to the set of dual quaternions $\mathbb{H}_{d}=\{Q_{d} : Q_{d} = Q + \epsilon Q_{\epsilon},~Q,Q_{\epsilon} \in \mathbb{H}\}$. Comparable to dual numbers, addition of dual quaternions $P_{d}$ and $Q_{d}$ is defined as an element wise addition
\begin{equation}
	P_{d}+Q_{d} = (P + Q) + \epsilon(P_{\epsilon} + Q_{\epsilon})
\end{equation}
and multiplication as
\begin{equation}
P_{d}Q_{d} = (PQ) + \epsilon(PQ_{\epsilon} + P_{\epsilon}Q) .
\end{equation}
Similar to the single quaternions, also the dual quaternions can be conjugated by conjugating both, the real quaternion and the dual quaternion, using the standard quaternion conjugation from equation \eqref{equ:QuatConj}
\begin{equation}
	Q_{d}^{*} = Q^{*} + Q_{\epsilon}^{*} 
	= q_{0} - q_{1}i - q_{2}j - q_{3}k + \epsilon( q_{\epsilon 0} - q_{\epsilon 1}i - q_{\epsilon 2}j - q_{\epsilon 3}k)
\end{equation}
\cite{N.Filipe.2013}. Furthermore, there exists a dual quaternion dual conjugate where also the dual number conjugation is incorporated \cite{Leclercq.2013}:
\begin{equation}
\bar{Q_{d}}^{*} = Q^{*} - Q_{\epsilon}^{*} 
= q_{0} - q_{1}i - q_{2}j - q_{3}k + \epsilon(- q_{\epsilon 0} + q_{\epsilon 1}i + q_{\epsilon 2}j + q_{\epsilon 3}k)
\end{equation}
The norm $\left\lVert Q_{d}\right\rVert$ of a dual quaternion can be calculated with
\begin{equation}
	\left\lVert Q_{d}\right\rVert = \sqrt{Q_{d}Q_{d}^{*}} = \sqrt{Q_{d}^{*}Q_{d}} 
	= \sqrt{(Q\cdot Q) + 2\epsilon (Q \cdot Q_{\epsilon})	}
\label{equ:dqNorm}
\end{equation}
Note that this norm is usually a dual number and no scalar value. \cite{N.Filipe.2013} 

Dual quaternions with $\left\lVert Q_{d}\right\rVert = 1$ are called unit dual quaternion and form the set $\mathbb{H}_{d}^{1}=\{Q_{d} : \left\|Q_{d}\right\| = 1,~Q_{d} \in \mathbb{H}_{d}\}$. Equation \eqref{equ:dqNorm} directly leads to two constrains for these dual unit quaternions: the real quaternion has to be a unit quaternion and the real Quaternion $Q$ and the dual quaternion $Q_{\epsilon}$ have to be orthogonal to each other. To norm a dual quaternion $Q_{d} = Q + Q_{\epsilon}$ and enforce these constrains the following method can be used \cite[37]{diss:Valverde}:
\begin{equation}
\begin{split}
	Q&:=\frac{Q}{\left\lVert Q\right\rVert} \\
	Q_{\epsilon}&:=
	\left( I_{4\times 4}-\dfrac{QQ^{T}}{\left\lVert Q\right\rVert^{2}}Q_{\epsilon} \right)
\end{split}
\label{equ:dqNormalization}
\end{equation}
As a measurement of the similarity between two dual quaternions $P_{d}$ and $Q_{d}$, an error dual quaternion $E_{d}$ can be introduced:
\begin{equation}
	E_{d} = P^{*}Q = E + \frac{1}{2}\epsilon E(\textbf{t} - \textbf{t}_{D}) 
\label{equ:errorDQ}
\end{equation}
where $E$ represents the rotation and $(\textbf{t} - \textbf{t}_{D})$ the translation necessary to align the two dual quaternions. This form makes the error dual quaternion $E_{d}$ a unit dual quaternion that can be seen by comparison with the following equations \eqref{equ:dqRotTrans}, \eqref{equ:dqTransRot} and the calculations in \eqref{equ:ProofPoseIsUnit}. \cite{N.Filipe.2013}

The matrix notation introduced in equations \eqref{equ:QuatLMatrix} - \eqref{equ:QuatConjMatrix}  can be extended for dual quaternions: \cite{Valverde.2018}

\begin{equation}
	[[P_{d}]]_{L}
	=
	\left[\begin{array}{cc}
	[[P]]_{L} & 0_{4\times 4} \\
	{[[P_{\epsilon}]]_{L}} & [[P]]_{L} \\
	\end{array}\right],~  
	[[Q_{d}]]_{R}
	=
	\left[\begin{array}{cc}
	[[Q]]_{R} & 0_{4\times 4} \\
	{[[Q_{\epsilon}]]_{R}} & [[Q]]_{R} \\
	\end{array}\right]
\label{equ:DQmatrix}	
\end{equation} 
\begin{equation}
	[[P_{d}^{*}]]_{L}
	=
	\left[\begin{array}{cc}
	[[P]]_{L}^{*} & 0_{4\times 4} \\
	{[[P_{\epsilon}]]_{L}}^{*} & [[P]]_{L}^{*} \\
	\end{array}\right],~  
	[[Q_{d}^{*}]]_{R}
	=
	\left[\begin{array}{cc}
	[[Q]]_{R} & 0_{4\times 4} \\
	{[[Q_{\epsilon}]]_{R}} & [[Q]]_{R} \\
	\end{array}\right]
\label{equ:DQconjMatrix}	
\end{equation} 
\begin{equation}
	[[\bar{P_{d}}^{*}]]_{L}
	=
	\left[\begin{array}{cc}
	[[P]]_{L}^{*} & 0_{4\times 4} \\
	{-[[P_{\epsilon}]]_{L}}^{*} & [[P]]_{L}^{*} \\
	\end{array}\right],~  
	[[\bar{Q_{d}^{*}}]]_{R}
	=
	\left[\begin{array}{cc}
	[[Q]]_{R}^{*} & 0_{4\times 4} \\
	{-[[Q_{\epsilon}]]_{R}}^{*} & [[Q]]_{R}^{*} \\
	\end{array}\right] .
\label{equ:DQdualConjMatrix}	
\end{equation} 
%
%
%Furthermore, a exponential for a dual quaternion $Q_{d} = Q + \epsilon Q_{\epsilon}$ exists also \cite[25]{BrunoVilhenaAdorno.2011}:
%\begin{equation}
%	exp(Q_{d}) = exp(Q) + \epsilon (Q_{\epsilon}exp(Q))
%\end{equation}
%%
Furthermore, there also exists an exponential for a dual quaternion $Q_{d} = Q + \epsilon Q_{\epsilon}$ \cite{NeilDantam.2014}:
\begin{equation}
	exp(Q_{d}) = e^{q_{0} + q_{\epsilon, 0}}
	\left(
		\cos(\phi) + 
		\frac{\sin(\phi)}{\phi} \textbf{q}
		+ \epsilon
		\left(
			\frac{-\sin(\phi)}{\phi} m + \frac{\sin(\phi)}{\phi} \textbf{q}_{\epsilon} + 
			\frac{\cos(\phi) - \frac{\sin(\phi)}{\phi}}{\phi^{2}}m\textbf{q}
		\right)
	\right)
\end{equation}
where
\begin{equation}
	\phi = \left\lVert \textbf{q} \right\rVert
\end{equation}
and
\begin{equation}
	m = \textbf{q} \cdot \textbf{q}_{\epsilon}.
\end{equation}
To avoid singularities for $\phi = 0$ the already known Taylor series from \eqref{equ:singularityQuatExpTailorSeries} and the following Taylor series can be used:
\begin{equation}
	\frac{\cos(\phi) - \frac{\sin(\phi)}{\phi}}{\phi^{2}}
	=
	- \frac{1}{3} + \frac{\phi^{2}}{30} - \frac{\phi^{4}}{840} + \frac{\phi^{6}}{45360} + \dots~.
\end{equation}
To calculate the logarithm $L_{d} = \ln(Q_{d}) = \ln(Q + Q_{\epsilon})$ of a dual quaternion, three intermediate results are used to simplify the calculation:
\begin{equation}
	\phi = \atantwo(\frac{\left\lVert \textbf{q}\right\rVert}{q_{0}}),
\end{equation}
\begin{equation}
	m = \textbf{q} \cdot \textbf{q}_{\epsilon},
\end{equation}
\begin{equation}
	\alpha = 
	\frac{q_{0} -
		\frac{\phi}{\left\lVert \textbf{q}\right\rVert}
		\left\lVert Q\right\rVert^{2}
	}
	{\left\lVert \textbf{q}\right\rVert^{2}}.
\end{equation}
With this, the real part of $L_{d}$ can be calculated with
\begin{equation}
	L = \ln(\left\lVert Q\right\rVert^{2}) + 
	\epsilon \frac{\phi}{\left\lVert \textbf{q}\right\rVert} \textbf{q}
\label{equ:DqLogReal}
\end{equation}
and the dual part with
\begin{equation}
	L_{\epsilon} = \frac{m + q_{0}q_{\epsilon,0}}
	{\left\lVert Q\right\rVert^{2}} + 
	\epsilon 
	\frac{m \alpha - q_{\epsilon, 0}}{\left\lVert Q\right\rVert^{2}} \textbf{q} +
	\frac{\phi}{\left\lVert \textbf{q}\right\rVert}\textbf{q}_{\epsilon}.
\label{equ:DqLogDual}
\end{equation}
Just like with the quaternion logarithm it becomes problematic when  $\left\| \textbf{q} \right\| \rightarrow 0$ while computing $\frac{\phi}{\left\lVert \textbf{q}\right\rVert}$. This can be handled using the already known method from equations \ref{equ:singularityDuatLn} and \ref{equ:singularityDuatLnTailorSeries}. Furthermore $\alpha$ can be rewritten as
\begin{equation}
	\alpha = 
	\frac{1}{\left\lVert Q\right\rVert} \left(
	\frac{\cos(\phi)}{\sin^{2}(\phi)}
	-
	\frac{\phi}{\sin^{3}(\phi)}
	\right)
\end{equation}
such that the Taylor series
\begin{equation}
	\frac{\cos(\phi)}{\sin^{2}(\phi)}
	-
	\frac{\phi}{\sin^{3}(\phi)}
	=
	-\frac{2}{3} - \frac{1}{5}\phi^{2} - \frac{17}{420} \phi^{4} - \frac{29}{4200} \phi^{6} + \dots
\end{equation}
can be used to calculate the final result. \cite{NeilDantam.2014}

\subsection{Rigid Transformation using Dual Quaternions}
\label{subsec:rbtDualQuaternions}

Similar to rotation quaternions like in equation \eqref{equ:rotQuat}, dual quaternions can also describe a rotation, but in addition they include a simultaneous translation. For this, the rotation quaternion $R \in \mathbb{H}_{1}$ and the pure quaternion $T=(0,\textbf{t})= t_{1}i+t_{2}j+t_{3}k$ form the dual quaternion
\begin{equation}
	Q_{d} = R + \dfrac{\epsilon}{2}TR .
\label{equ:dqRotTrans}
\end{equation}
Using a point P as a pure quaternion $P = (0,\textbf{p}) = p_{1}i + p_{2}j + p_{3}k$ and extending it to a dual quaternion $P_{d} = 1 + \epsilon P$, the rotation and translation of $P_{d}$ with $Q_{d}$ to the new position $P_{d}'$ can be described as the following:
\begin{equation}
	\begin{split}	
		P_{d}' &= Q_{d}P_{d}\bar Q_{d}^{*} \\
	  	&= (R + \dfrac{\epsilon}{2}TR)(1 + \epsilon P)(R^{*} - \dfrac{\epsilon}{2}(TR)^{*}) \\
	  	&= (R + \dfrac{\epsilon}{2}TR + \epsilon RP)(R^{*} - \dfrac{\epsilon}{2}R^{*}T^{*}) \\
	  	&= RR^{*} + \epsilon(\dfrac{1}{2} (TRR^{*}-RR^{*}T^{*}) + RPR^{*}) \\
	  	&= 1 + \epsilon(\dfrac{1}{2} (T-T^{*}) + RPR^{*}) \\
	  	&= 1 + \epsilon(RPR^{*} + T)
	\end{split}
\end{equation}
Here, $(RTR^{*})$ describes a rotated point after \eqref{equ:RotatingWithSingleQuaternions}, implying that the order of sequence is first rotation, afterwards translation. Also, note that in the second last line the property $-Q = Q^{*}$ of pure quaternions is utilized.

\begin{figure}[H]
	\begin{subfigure}{\textwidth}
		\centering
		\includegraphics[width=0.9\linewidth]{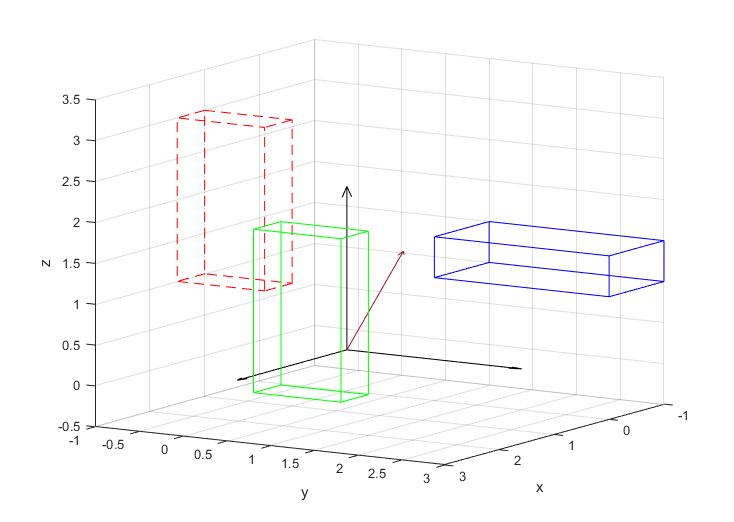}
		%\caption{Output $y_{1}$ - $r(t)$}
		\label{fig:rbtExampleXYZ}
	\end{subfigure}
	\begin{subfigure}{0.33\textwidth}
		\centering
		\includegraphics[width=1\linewidth]{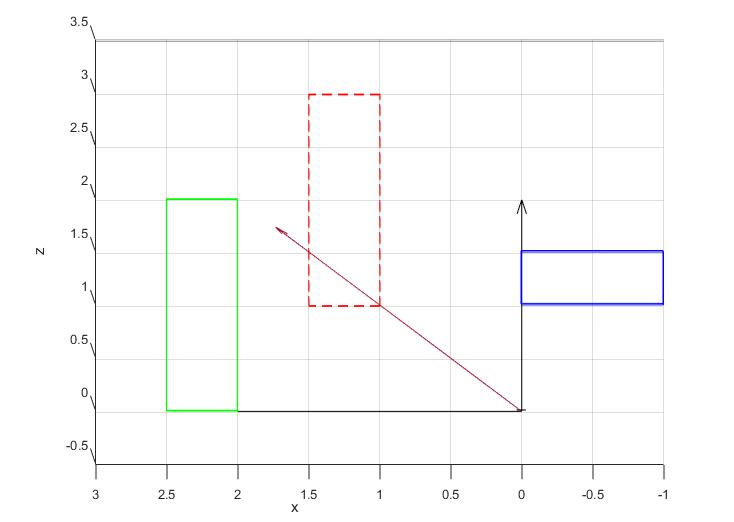}
		%\caption{Output $y_{1}$ - $r(t)$}
		\label{fig:rbtExampleXZ}
	\end{subfigure}
	\begin{subfigure}{0.33\textwidth}
		\centering
		\includegraphics[width=1\linewidth]{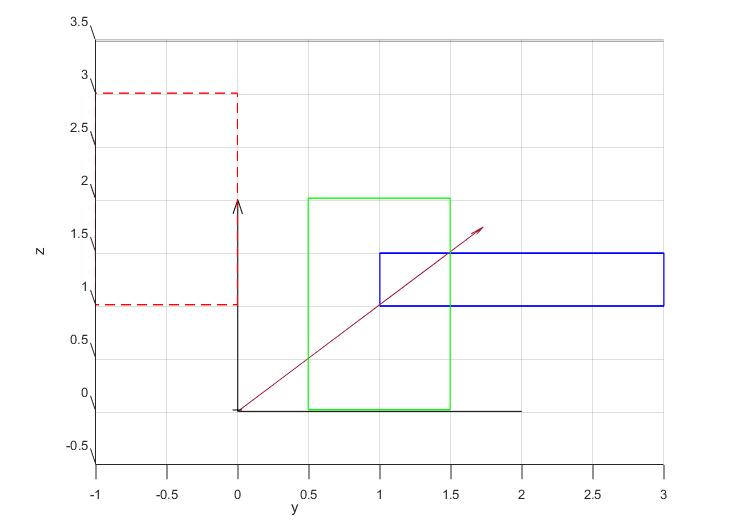}
		%\caption{Output $y_{2}$ - $\phi(t)$}
		\label{fig:rbtExampleYZ}
	\end{subfigure}
	\begin{subfigure}{0.33\textwidth}
		\centering
		\includegraphics[width=1\linewidth]{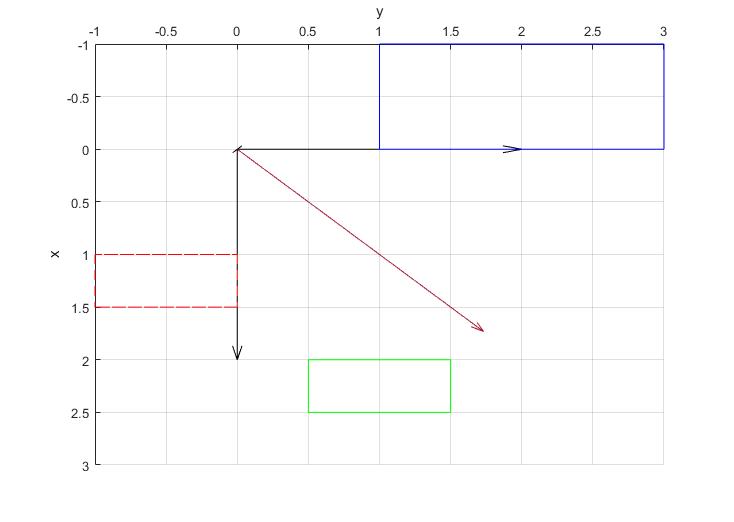}
		%\caption{Output $y_{2}$ - $\phi(t)$}
		\label{fig:rbtExampleXY}
	\end{subfigure}
	\caption{Example for a rigid body transformation. In blue the initial position, in red the position after the rotation, in green the final position after rotation and translation. The arrow indicates the rotation axis.}
	\label{fig:rbtExample}
\end{figure}

Selecting the transformation dual quaternion as described in \eqref{equ:dqRotTrans} always results in a unit dual quaternion:
\begin{equation}
\begin{split}	
	Q_{d}Q_{d}^{*} &= (R + \dfrac{\epsilon}{2}TR)(R^{*} + \dfrac{\epsilon}{2}(TR)^{*}) \\
	&= (R + \dfrac{\epsilon}{2}TR)(R^{*} + \dfrac{\epsilon}{2}R^{*}T^{*}) \\
	&= RR^{*} + \epsilon(\dfrac{1}{2} (RR^{*}T^{*}+TRR^{*}) + RTR^{*}) \\
	&= 1 + \epsilon \dfrac{1}{2} (T+T^{*}) \\
	&= 1
	\end{split}
\label{equ:ProofPoseIsUnit}
\end{equation}
However, selecting the transformation dual quaternion $Q_{d}$ as 
\begin{equation}
	Q_{d} = R + \dfrac{\epsilon}{2}RT
\label{equ:dqTransRot}
\end{equation}
changes the order of sequence to first translation and afterwards rotation. 
\begin{equation}
	\begin{split}	
		P_{d}' &= Q_{d}P_{d}\bar Q_{d}^{*} \\
		&= (R + \dfrac{\epsilon}{2}RT)(1 + \epsilon P)(R - \dfrac{\epsilon}{2}(RT)^{*}) \\
		&= (R + \dfrac{\epsilon}{2}RT + \epsilon RP)(R^{*} - \dfrac{\epsilon}{2}T^{*}R^{*}) \\
		&= RR^{*} + \epsilon(\dfrac{1}{2} (RTR^{*}-RT^{*}R^{*}) + RPR^{*}) \\
		&= 1 + \epsilon((RTR^{*}) + RPR^{*}) \\
		&= 1 + \epsilon R(P + T)R^{*}
	\end{split}
\end{equation}
This time the already translated point $(P+T)$ is rotated with the rotation quaternion $R$. Of course the unit property holds for this dual quaternion as well and can be calculated similarly.
\\
\\
For a dual quaternion $Q_{d} = Q + \epsilon Q_{\epsilon}$ in the form of \eqref{equ:dqRotTrans}, the translation can be extracted with
\begin{equation}
	T = 2 Q_{\epsilon}Q^{*}
\label{equ:extractTrans}
\end{equation}
and for \eqref{equ:dqTransRot} with
\begin{equation}
	T = 2 Q^{*} Q_{\epsilon}
\label{equ:extractTrans2}
\end{equation}
\\
\\
An example for a rigid body transformation is shown in figure \ref{fig:rbtExample}. The rotation axis there is $\textbf{n}=(1,1,1)$, the rotation angle $\theta =90^{\circ}$ and the translation $(1, 1.5, -1)$. Equation \eqref{equ:dqRotTrans}, and therefore first rotation and then translation, was used. The cuboid rigid bodies were described with points indicating each corner. Therefore, per rigid body, eight transformations were done. 

\noindent
It might seem tempting to describe a rigid body using its centre of mass and a quaternion for the orientation, encoded as shown in \eqref{equ:dqRotTrans} or \eqref{equ:dqTransRot} and doing the same transformation as introduced for points.  However, this is not possible which can be shown with the following calculations where the rigid body is described as $B_{d} = O + \frac{\epsilon}{2}PO$:
\begin{equation}
\begin{split}	
	\hat B_{d} &= Q_{d}B_{d}\bar Q_{d}^{*} \\
	&= (R + \frac{\epsilon}{2}TR)(O + \frac{\epsilon}{2} PO)(R^{*} - \frac{\epsilon}{2}(TR)^{*}) \\
	&= (RO + \frac{\epsilon}{2}RPO + \frac{\epsilon}{2} TRO)(R^{*} - \frac{\epsilon}{2}R^{*}T^{*}) \\
	&= ROR^{*} - \frac{\epsilon}{2}ROR^{*}T^{*} + \frac{\epsilon}{2} RPOR^{*} + \frac{\epsilon}{2}TROR^{*} \\
	&= ROR^{*} + \frac{\epsilon}{2}(-ROR^{*}T^{*} + RPOR^{*} + TROR^{*}) \\
	&=R' + \frac{\epsilon}{2}(-R'T^{*} + RPOR^{*} + TR')
\end{split}
\end{equation}
Analysing the result leads to the finding that the orientation quaternion is rotated correctly (denoted as $R'$) but also incorporated in the translation. Furthermore, the orientation quaternion $O$ is included in the term describing the rotated point which is also incorrect. A similar calculation can be done for the other possible sequence of actions with first translation and afterwards rotation. 
 
Therefore, if a rigid body shall be described using \eqref{equ:dqRotTrans} or \eqref{equ:dqTransRot}, another transformation approach has to be used, which is further described in subsection \ref{subsec:RbKinematicsUsingDq}.

\section{Physical Preliminaries}
After introducing the necessary mathematical preliminaries, now the physical preliminaries are targeted. These are in particular rigid body kinematics, rigid body rotational kinematics with the usage of quaternions and finally rigid body kinematics with dual quaternions, where the rotational movement is extended with translational movement.

\subsection{Rigid Body Kinematics}
\label{subsec:rigidBodyKinematics}

The rotational movement of a rigid body in an inertial system, i.e. a system where no forces act on the body, with its origin in the centre of mass and not rotated to the world coordinate system can be described with 
\begin{equation}
	L = \textbf{I}\omega
\label{equ:AngMomTensorOmega}
\end{equation}
where $L$ is the body's angular momentum, $\textbf{I}$ its moment of inertia tensor and $\omega$ the angular velocity.

$\textbf{I}$ is a two-dimensional tensor with the general form
\begin{equation}
	\textbf{I} = 
	\left[\begin{array}{cccc}
	I_{11} & I_{12} & I_{13}  \\
	I_{21} & I_{22} & I_{23} \\	
	I_{31} & I_{32} & I_{33} \\
	\end{array}\right]
\end{equation}
where the diagonal elements are called moments of inertia and the secondary diagonal elements moments of deviation.
This leads to an alternative representation of equation \eqref{equ:AngMomTensorOmega}:
\begin{equation}
	\begin{split}	
		L_{1} &= I_{11} \omega_{x} + I_{12} \omega_{y} + I_{13} \omega_{z} \\
		L_{2} &= I_{21} \omega_{x} + I_{22} \omega_{y} + I_{23} \omega_{z} \\
		L_{3} &= I_{31} \omega_{x} + I_{32} \omega_{y} + I_{33} \omega_{z} \\
	\end{split}
\label{equ:AngMomTensorOmega2}
\end{equation}
Hence, the rigid body's angular momentum and angular velocity are not parallel in the general case.
\cite[131-133]{Bartelmann.2018}

It has to be pointed out that both, the moment of inertia tensor and the angular velocity, are generally time dependent such that \eqref{equ:AngMomTensorOmega} should be written as
\begin{equation}
	L = \textbf{I}(t)\omega(t)
\end{equation}
but for the sake of simplicity the time dependency is omitted in this section when not explicitly needed. \cite[137-139]{Bartelmann.2018} Since an inertial system is considered, the angular momentum $L$ is however constant.

This time dependency can be eliminated by fixing a coordinate system to the rigid body and therefore letting it rotate with the body. Furthermore, equation \eqref{equ:AngMomTensorOmega2} can be simplified by rotating it 
such that $\textbf{I}$ has diagonal form. This is possible due to the fact that the moment of inertia tensor is symmetric. The subscript $body$ denotes this special body fixed coordinate system.
%
%This time dependency can be eliminated by choosing a special coordinate system, where the subscript $body$ denotes this body fixed coordinate system: initially, it has to be fixed to the rigid body and therefore rotates with it. Subsequently, it is rotated such that $\textbf{I}$ has diagonal form, which is possible due to the fact that the moment of inertia tensor is symmetric.
This yields
\begin{equation}
	\textbf{I}_{body} = 
	R^{T}\textbf{I}R
	= 
	\left[\begin{array}{cccc}
		I_{1} & 0 & 0 \\
		0 & I_{2} & 0 \\	
		0 & 0 & I_{3} \\
	\end{array}\right]
\end{equation}
with $R$ as the appropriate rotation matrix. These coordinate system axes are called principal axes of inertia. \cite[139]{Bartelmann.2018}

Hereby equation \eqref{equ:AngMomTensorOmega2} simplifies to 
\begin{equation}
	\begin{split}	
	L_{1} &= I_{1} \omega_{x} \\
	L_{2} &= I_{2} \omega_{y} \\
	L_{3} &= I_{3} \omega_{z} . \\
	\end{split}
\end{equation}

\subsection{Rigid Body Rotational Kinematics using Quaternions}

Rotating a rigid body with an angular velocity $\omega$ leads to a change in the quaternion describing its attitude. This relationship can be expressed with the equation
\begin{equation}
\dot Q = \dfrac{1}{2} \omega_{world} Q= \dfrac{1}{2} Q\omega_{body} 
\end{equation}
where it has to be taken care of the angular velocity's frame of reference. \cite{N.Filipe.2013}

A solution for this first order differential equation is given with
\begin{equation}
	Q(t) = exp(\frac{t}{2}\omega)Q(0)
\label{equ:solutionFirstOrderDiff}
\end{equation}
where, similar to regular first order differential equations, the exponential is used. However, this is only a viable solution for a constant angular velocity $\omega$. \cite{Andrle.2013} 

When this is not the case, another solution must be found. 
A simple approach to solve this first order differential equation with a varying $\omega$ could be done with the Euler method, using 
\begin{equation}
	Q(t + \Delta t) = Q(t) + \dot {Q} \Delta t = Q(t) + \dfrac{1}{2} \omega_{world}(t) Q(t) \Delta t
\end{equation}
and an initial orientation $Q(0)$. However, due to the addition, the unit property of the quaternion describing a rotation can't be ensured, therefore a re-normalization is necessary after each integration step $\Delta t$, which could cause further numerical errors.  
 
More advanced integration methods like the Leap-frog method \cite{Zhao.2013} or the Runge-Kutta-Method \cite{Andrle.2013} are available for quaternions, but they suffer from the same problem since the unit property of quaternions can't be preserved under addition.

Therefore, a solution comparable to \eqref{equ:solutionFirstOrderDiff}, but with a changing $\omega(t)$ might be favourable. One approach could be the s-stage Crouch-Grossman algorithm, with an exemplary three stage implementation for quaternions like shown in the following \cite{Andrle.2013}:
\begin{equation}
\begin{split}	
	Q(t + \Delta t) &= exp(\frac{24}{17} \Delta t K^{(3)}) exp(\frac{-2}{3} \Delta t K^{(2)}) exp(\frac{13}{51} \Delta t K^{(1)}) Q(t) \\
	K^{(1)} &= \frac{1}{2} \omega(t) \\
	K^{(2)} &= \frac{1}{2} \omega(t + \frac{3}{4}\Delta t) \\
	K^{(3)} &= \frac{1}{2} \omega(t + \frac{17}{24}\Delta t) .\\
\end{split}	
\end{equation}
However, this method has the limitation that $\omega(t)$ as a function of time has to be known to calculate the $K^{(i)}$ values. The values at the integration steps alone are not sufficient since intermediate values for the angular velocity are required.  

Another approach, combining the preservation of the unit property from \eqref{equ:solutionFirstOrderDiff} and simplicity of the Euler method is the following \cite{NeilDantam.2014}:
\begin{equation}
Q_{d}(t + \Delta t) = exp(\frac{\Delta t}{2}\omega(t)) Q_{d}(t)
\end{equation}
This allows the integration of orientations as quaternions with given values for the angular velocity at discrete time steps $\Delta t$ under the assumption that the change in the angular velocity in this time step is rather small or otherwise that the time step is small enough to keep the error low.

\subsection{Rigid Body Kinematics using Dual Quaternions}
\label{subsec:RbKinematicsUsingDq}

Similar to the rotation in quaternion space, also for the combination of rotation and translation in dual quaternion space a description for the kinematics and therefore a change in the combined orientation and location dual quaternion exists \cites{Silva.2018}{Zhang.2010}:

\begin{equation}
	\dot Q_{d} = \dfrac{1}{2} \xi_{world} Q_{d} = \dfrac{1}{2} Q_{d} \xi_{body}
    \label{equ:DqDiffEquation}
\end{equation}
where 
\begin{equation}
\xi_{world} = \omega_{world} + \epsilon(v_{world} + p_{world} \times \omega_{world}),~ \xi_{body} = \omega_{body} + \epsilon(v_{body} + \omega_{body} \times p_{body}) .
\label{equ:DqTwist}
\end{equation}
This is often referred to as the twist of the rigid body \cites{Wu.2005}{Zhang.2010}.

Here, the solution of this first order differential equation is also available with the exponential of the this time dual quaternion
\begin{equation}
	Q_{d}(t + \Delta t) = exp(\frac{\Delta t}{2}\xi) Q_{d}(t)
\label{equ:DqIntegration}
\end{equation}
as a numerical integration with the discrete time step $\Delta t$ and a given initial pose $Q_{d}(0)$ \cite{Savino.2020}.

\section{Neural Network Architectures in Dual Quaternion Space}

\subsection{Dual Quaternion Feed Forward Neural Networks}

The concept of a FFN can be adapted to operate in the dual quaternion space as well. In this case, each neuron, weight and bias becomes a dual quaternion instead of a scalar value. Again, a general description for this net can be formulated:
\begin{equation}
	x^{(l+1)} = \psi (z^{(l+1)}) = \psi (W^{(l)}x^{(l)} + b^{(l)})
\label{equ:dqnnFfnGeneral}
\end{equation} 
with $x^{(l+1)} \in \mathbb{H}_{d}$ as the output of layer $l$ and the input for layer $l+1$, the dual quaternion activation function $\psi(\cdot)$, $W \in \mathbb{H}_{d}$ as the dual quaternion weights, $x^{(l)} \in \mathbb{H}_{d}$ as the input of layer $l$ and $b \in \mathbb{H}_{d}$ as the dual quaternion bias.

The output $Y_{d,i}$ of a single dual quaternion neuron $i$ with $R$ inputs is
\begin{equation}
	Y_{d,i} = \psi (Z_{d,i}), Z_{d,i} = \sum_{i=1}^{R} W_{d,i}X_{d,i} + B_{d,i} .
\label{equ:dqSingleNeuronSum}
\end{equation}
For efficient calculation, the convenient matrix/vector notation from the FFN's can be adapted also. For this, the matrix notation of dual quaternions can be utilised. The matrix/vector representation of equation \eqref{equ:dqSingleNeuronSum} is
\begin{equation}
	Z_{d,i} = \left[\begin{array}{cccc}
			[[W_{d,1}]]_{L} & [[W_{d,2}]]_{L} & \cdots & [[W_{d,R}]]_{L} 
		\end{array}\right]
		\left[\begin{array}{cc}
			X_{1} \\ X_{2} \\ \vdots \\ X_{R}
		\end{array}\right]
		+
		B_{i}
\end{equation}
where $X_{i}$ is the vector representation of the dual quaternion $X_{d,i}$. The same calculation has an alternative representation in
\begin{equation}
	Z_{d,i} = \left[\begin{array}{cccc}
		[[X_{d,1}]]_{R} & [[X_{d,2}]]_{R} & \cdots & [[X_{d,R}]]_{R} 
	\end{array}\right]
	\left[\begin{array}{c}
		W_{1} \\ W_{2} \\ \vdots \\ W_{R}
	\end{array}\right]
	+
	B_{i} .
\end{equation}
Several possibilities exist to extend this notation to calculate the whole layer in a single matrix calculation, not least because of the two different dual quaternion neuron formulas. For a layer with $R$ inputs and $S$ neurons one method is the following:
\begin{equation}
	Z_{d}^{g} = \left[\begin{array}{cccc}
		[[X_{d,1}]]_{R} [[X_{d,2}]]_{R} \cdots [[X_{d,R}]]_{R}) 
	\end{array}\right]
	\begingroup
	\setlength\arraycolsep{2pt}
	\left[\begin{array}{cccc}
		W_{1,1}  & W_{1,2} & \cdots & W_{1,S}\\ 
		W_{2,1}  & W_{2,2} & \cdots & W_{1,S}\\ 
		\vdots  & \vdots & \ddots & \vdots\\ 
		W_{R,1} & W_{R,2} & \cdots & W_{R,S} \\
	\end{array}\right]
	+
	\left[\begin{array}{cccc}
		B_{1,1} & B_{1,2} & \cdots & B_{1,S}\\ 
	\end{array}\right]
	\endgroup
\label{equ:dqMatrixVar1}
\end{equation}
where the input matrix has the dimension $8 \times 8R$, and the weights and bias matrices the dimensions $8R \times S$ and $8 \times S$, resulting in a $8 \times S$ output.

Another approach is
\begin{equation}
	Z_{d}^{g} = 
	\left[\begin{array}{cccc}
		[[W_{d,1,1}]]_{L} & [[W_{d,1,2}]]_{L} & \cdots & [[W_{d,1,R}]]_{L} \\
 		{[[W_{d,2,1}]]_{L}} & [[W_{d,2,2}]]_{L} & \cdots & [[W_{d,2,R}]]_{L} \\
		\vdots & \vdots & \ddots & \vdots \\
		{[[W_{d,S,1}]]_{L}} & [[W_{d,S,2}]]_{L} & \cdots & [[W_{d,S,R}]]_{L} \\
	\end{array}\right]
	\left[\begin{array}{cc}
		X_{1} \\ X_{2} \\ \vdots \\ X_{R}
	\end{array}\right]
	+
	\left[\begin{array}{cc}
		B_{1} \\ B_{2} \\ \vdots \\ B_{S}
	\end{array}\right]
\label{equ:dqMatrixVar2}	
\end{equation}
where the weight matrix has the dimension $8S \times 8R$, and input and bias have the dimension $8R \times 1$ respectively $8S \times 1$ yielding an output of dimension $8S \times 1$.

The results of both methods are mathematically identical except for the output dimensions, it depends on the implementation which method is the favourable.
\\
\\
Since this model works based on additions, the network's outputs are usually not unit dual quaternions. If this is desired for the output, either normalisation with equation \eqref{equ:dqNormalization} or a network based on multiplication is a possibility. The latter however is only possible if all inputs are unit dual quaternions and all weights as well. This requires a randomized unit initialization and it has to be taken care that all dual quaternion weights stay unit during optimization with backpropagation. Furthermore, special dual quaternion activation functions $\sigma(\cdot)$, which are also unit preserving, have to be used. All this yields the alternative formulation for the output $x_{i}$ of a single dual quaternion neuron:
\begin{equation}
Y_{d,i} = \sigma (Z_{d,i}), Z_{d,i} = \prod_{i=1}^{R} W_{d,i}X_{d,i}.
\label{equ:dqSingleNeuronProd}
\end{equation}
A unit dual quaternion weight $W_{d} \in \mathbb{H}_{d}^{1}$  can be realised with 
\begin{equation}
	W_{d} = W + \epsilon W_{\epsilon},~
	W = \left[\begin{array}{c}
			\sqrt{1 - \left\lVert \mathbf{\Phi} \right\rVert}  \\
			{\mathbf{\Phi}}
		\end{array}\right],~
	W_{\epsilon} = \left[
		\begin{array}{c}
			-\frac{\mathbf{\Psi\Phi}}{\sqrt{1 - \left\lVert \mathbf{\Phi} \right\rVert}} \\
			\mathbf{\Psi}
		\end{array}
	\right]
\end{equation}
where $\mathbf{\Phi}$ and $\mathbf{\Psi}$ are randomly initialized three-element-vectors with the constraint that $\left\lVert\Phi\right\rVert \leq 1$.

\subsection{Dual Quaternion Attention}
\label{sec:DqAttention}

The concepts and ideas of attention are also applicable in dual quaternion space, however some adjustments have to be made to respect the properties and meanings of dual quaternions.

\begin{figure}[h!]
	\centering
	\includegraphics[width=0.90\linewidth]{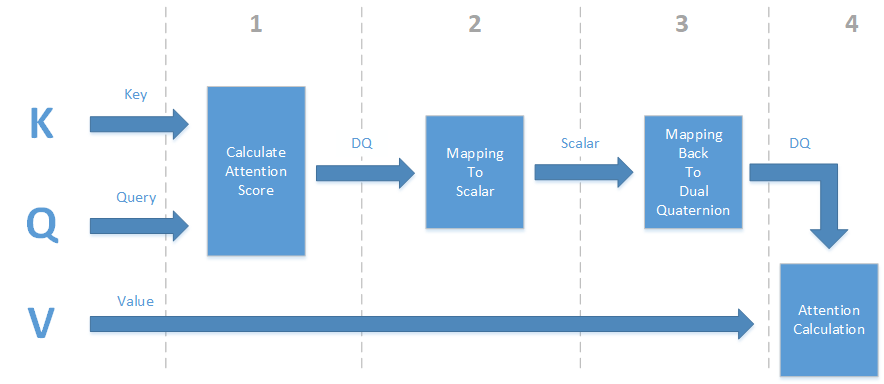}
	\caption{The four stages necessary to apply the attention mechanism in dual quaternion space}
	\label{fig:attentionStages}
\end{figure}

The whole process can be divided into four main stages as shown in figure \ref{fig:attentionStages}: calculating the attention scores in dual quaternion space, mapping them to a scalar attention value, bringing this scalar back to the dual quaternion space and finally multiplication with the input values.
These stages and possible variations are further described in the following. Here, the very general form with keys, queries and values is assumed. However just because they can be used, they don't need to, the following calculations are applicable without limitations if keys or queries are not existent. Depending on the application, the keys $K$ and/or queries $Q$ can be equal to the values $V$.

\subsubsection{Attention Score Calculation}

To calculate dual quaternion attention scores, attention strategies from regular neural networks are now brought into the dual quaternion space.

\paragraph{Dual Quaternion Additive Attention}
\label{subsubsec:dqAdditiveAttention}

Additive attention in dual quaternion space can be realised straight forward using equation \eqref{equ:dqnnFfnGeneral} and equation \eqref{equ:dqMatrixVar1} respectively \eqref{equ:dqMatrixVar2} without the usage of a bias: 
\begin{equation}
	A_{d}^{g} = \mathbf{W}_{d}^{g^2}(\mathbf{W}_{d}^{g^1} X_{d}^{g}) .
\end{equation}
For the case that the keys and queries are not equal, they have to be concatenated to form the input dual quaternion vector $X_{d}^{g} = [K_{d}^{g}, Q_{d}^{g}]^{T}$, otherwise it is sufficient to use only the keys as input vector $X_{d}^{g}$.

%However, using the element-wise working $softmax()$ on the resulting dual quaternions is problematic since they can't be treated like regular scalar values as in the regular case. Furthermore it has to be taken great care of how this result is used, because it is a dual quaternion and no scalar which can be used as a weight. This leaves room for several strategies, one could be for example omitting the $softmax()$, define a target dual quaternion and use the output dual quaternions to calculate a similarity to this desired target with e.g. the cosine-similarity or the mean-square error. This results in a mapping from the dual quaternion space to scalar values, and now the $softmax()$ or any other function could be used to define a weight $\lambda$, which is afterwards used to rate the dual quaternion inputs $X_{i}$ through multiplication with $ \lambda(1 + 0i +0j + 0k + \epsilon0)$. Likewise, any other mapping from dual quaternion space to a scalar value would work for this strategy. Another approach could be to do a direct multiplication of the input quaternions with the attention output quaternions, resulting in an initial transformation of the inputs before being fed in the following layers. The right approach can't be generalized since it is too much dependent on the actual task which the neural network shall fulfil. 

\paragraph{Dual Quaternion Multiplicative Attention}

The dual quaternion space limits the direct portability of the multiplicative attention, because the dot-product of two dual quaternions does not have a useful meaning, hence it has to be adapted. One possible approach is utilising the dual quaternion property $Q_{d}Q_{d}^{*}=1$, to calculate a similarity quaternion or error quaternion between all inputs. This can be done through element-wise multiplication as shown in equation \eqref{equ:dqmulattention} :
%%
%\begin{equation}
%	A = X_{d}^{g} \circ X_{d}^{g^*}
%\end{equation}
%%
%
\begin{equation}
	A_{d}^{g} = K_{d}^{g} \circ Q_{d}^{*,g} .
\label{equ:dqmulattention}
\end{equation}
The conjugate of the query is important here in order to exploit the similarity property explained above since key and query can be the same. The straight forward matrix implementation of this is 
%%
%\begin{equation}
%	A = 
%	\left[\begin{array}{c}
%		[[X_{d,1}]]_{L} \\
%		{[[X_{d,2}]]_{L}} \\
%		\vdots \\
%		{[[X_{d,R}]]_{L}} \\
%	\end{array}\right]
%	\left[\begin{array}{cccc}
%		X_{1}^{*} & X_{2}^{*} & \cdots & X_{R}^{*}
%	\end{array}\right] .
%\label{equ:dqmulattention}	
%\end{equation}
%%
%
\begin{equation}
A_{d}^{g} = 
\left[\begin{array}{cccc}
[[K_{d,1}]]_{L} & 0_{8\times 8} & \cdots & 0_{8\times 8} \\
0_{8\times 8} & {[[K_{d,2}]]_{L}} & \cdots & 0_{8\times 8} \\
\vdots  & \vdots & \ddots & \vdots \\
0_{8\times 8} & \cdots & 0_{8\times 8} & {[[K_{d,R}]]_{L}} \\
\end{array}\right]
\left[\begin{array}{c}
Q_{1}^{*} \\
Q_{2}^{*} \\
\cdots \\
Q_{R}^{*}
\end{array}\right] .
\end{equation}
resulting in an $8R \times 1$ vector $A$. Likewise, the similarity of all dual quaternions to each other can be calculated with
\begin{equation}
\mathbf{A}_{d}^{g} = 
\left[\begin{array}{c}
[[K_{d,1}]]_{L} \\
{[[K_{d,2}]]_{L}} \\
\vdots \\
{[[K_{d,R}]]_{L}} \\
\end{array}\right]
\left[\begin{array}{cccc}
Q_{1}^{*} & Q_{2}^{*} & \cdots & Q_{R}^{*}
\end{array}\right] .
\label{equ:dqmulattentionKeyQuery}	
\end{equation}
For $R$ inputs, the input dimensions are $8R \times 8$ and $8 \times R$, resulting in a $8R \times R$ output in case of conjugates $Q_{i}^{*}$ in vector representation. When the conjugates are also used in matrix form $[[Q_{i}^{*}]]_{L}$, then the inputs of dimensions $8R \times 8$ and $8 \times 8R$ result in an $8R \times 8R$ similarity matrix. Because of the dual quaternion property $Q_{d}Q_{d}^{*} = Q_{d}^{*}Q_{d}$, this matrix is a symmetric matrix. In the case that keys and queries are the same, it contains ones on the main diagonal. 

Since the upper triangular matrix equals the lower one, it's sufficient to continue calculation with one of it because all information are included there.\\
\\
As an extension to this, also additional weights $W_{d}^{g^K}$ for the keys and $W_{d}^{g^Q}$ for the queries can be introduced to perform an initial transformation as shown in the following equation \eqref{equ:dqmulattentionWithWeights}:   
\begin{equation}
	\mathbf{A}_{d}^{g} 
	= 
	W_{d}^{g^K} K_{d}^{g} \left(W_{d}^{g^Q}Q_{d}^{g}\right)^{*} 
	=
	W_{d}^{g^K} K_{d}^{g}  Q_{d}^{*,g}  W_{d}^{*,g^Q}
\label{equ:dqmulattentionWithWeights}	
\end{equation}
The corresponding matrix-calculation for this operation is
\begin{equation}
\begin{scriptsize}
	\mathbf{A}_{d}^{g} = 
	\begingroup
	\setlength\arraycolsep{2pt}
	\left[\begin{array}{cccc}
		[[W_{d,1}^{K}]]_{L} & 0_{8\times 8} & \cdots & 0_{8\times 8} \\
		0_{8\times 8} & {[[W_{d,2}^{K}]]_{L}} & \cdots & 0_{8\times 8} \\
		\vdots  & \vdots & \ddots & \vdots \\
		0_{8\times 8} & \cdots & 0_{8\times 8} & {[[W_{d,R}^{K}]]_{L}} \\
	\end{array}\right]
	\left[\begin{array}{c}
		[[K_{d,1}]]_{L} \\
		{[[K_{d,2}]]_{L}} \\
		\vdots \\
		{[[K_{d,R}]]_{L}} \\
	\end{array}\right]
	\left[\begin{array}{c}
		[[Q_{d,1}^{*}]]_{L} \\
		{[[Q_{d,2}^{*}]]_{L}} \\
		\vdots \\
		{[[Q_{d,R}^{*}]]_{L}}
	\end{array}\right]^{T}
	\left[\begin{array}{cccc}
		W_{1}^{*^Q} & 0_{8\times 1} & \cdots & 0_{8\times 1} \\
		0_{8\times 1} & W_{2}^{*^Q} & \cdots & 0_{8\times 1} \\
		\vdots  & \vdots & \ddots & \vdots \\
		0_{8\times 1} & \cdots & 0_{8\times 1} & W_{R}^{*^Q} \\
	\end{array}\right]
	\endgroup
	.
\label{equ:dqmulattentionWithWeights2}	
\end{scriptsize}
\end{equation}
Here, the dual quaternion property $(W_{d}Q_{d})^{*} = Q_{d}^{*}W_{d}^{*}$ was utilized to ensure that the weights are multiplied from the same side in both cases.
The output is again of the size $8R \times R$. For an output of $8R \times 8R$ and hence dual quaternion matrix representations of the attention scores, the conjugate weights $W_{i}^{*^Q}$  need to be replaced with weights $[[W_{i}^{*^Q}]]_{L}$ in matrix form.
\\
\\
Instead of doing a multiplication with the weights from the left, this can also be done from the right with
%
%\begin{equation}
%	a_{i} = X_{d, i}W_{d, i} (X_{d, i}W_{d, i})* = X_{d, i}W_{d, i}W_{d, i}^{*}X_{d, i}^{*}
%\end{equation}
%
\begin{equation}
\mathbf{A}_{d}^{g}
= 
K_{d}^{g} W_{d}^{g^K} \left(Q_{d}^{g} W_{d}^{g^Q}\right)^{*} 
=
K_{d}^{g} W_{d}^{g^K}  W_{d}^{*,g^Q} Q_{d}^{*,g}
\end{equation}
\label{equ:dqmulattentionWithWeightsGedreht}	
The matrix-calculation of the overall attention scores $A$ then is
	\begin{equation}
	\begin{scriptsize}
	\mathbf{A}_{d}^{g} = 
	\begingroup
	\setlength\arraycolsep{2pt}
	\left[\begin{array}{cccc}
		[[K_{d,1}]]_{L} & 0_{8\times 8} & \cdots & 0_{8\times 8} \\
		0_{8\times 8} & {[[K_{d,2}]]_{L}} & \cdots & 0_{8\times 8} \\
		\vdots  & \vdots & \ddots & \vdots \\
		0_{8\times 8} & \cdots & 0_{8\times 8} & {[[K_{d,R}]]_{L}} \\
	\end{array}\right]
	\left[\begin{array}{c}
		[[W_{d,1}^{K}]]_{L} \\
		{[[W_{d,2}^{K}]]_{L}} \\
		\vdots \\
		{[[W_{d,R}^{K}]]_{L}} \\
	\end{array}\right]
	\left[\begin{array}{c}
		[[W_{d,1}^{*,Q}]]_{L} \\
		{[[W_{d,2}^{*,Q}]]_{L}} \\
		\vdots \\
		{[[W_{d,R}^{*,Q}]]_{L}}
	\end{array}\right]^{T}
	\left[\begin{array}{cccc}
		Q_{1}^{*} & 0_{8\times 1} & \cdots & 0_{8\times 1} \\
		0_{8\times 1} & Q_{2}^{*} & \cdots & 0_{8\times 1} \\
		\vdots  & \vdots & \ddots & \vdots \\
		0_{8\times 1} & \cdots & 0_{8\times 1} & Q_{R}^{*} \\
	\end{array}\right]
	\endgroup
	.
	\end{scriptsize}
	\label{equ:dqmulattentionWithWeights3}	
	\end{equation}
Since dual quaternion multiplication is associative, the two weight matrices can be combined to one single weight matrix with the shape $8R \times 8R$, resulting in the following equation:
	\begin{equation}
	\begin{tiny}
	\mathbf{A}_{d}^{g} = 
	\begingroup
	\setlength\arraycolsep{1pt}
	\left[\begin{array}{cccc}
		[[K_{d,1}]]_{L} & 0_{8\times 8} & \cdots & 0_{8\times 8} \\
		0_{8\times 8} & {[[K_{d,2}]]_{L}} & \cdots & 0_{8\times 8} \\
		\vdots  & \vdots & \ddots & \vdots \\
		0_{8\times 8} & \cdots & 0_{8\times 8} & {[[K_{d,R}]]_{L}} \\
	\end{array}\right]
	\left[\begin{array}{cccc}
		[[W_{d,1,1}]]_{L} & [[W_{d,1,2}]]_{L} & \cdots & [[W_{d,1,R}]]_{L} \\
		{[[W_{d,2,1}]]_{L}} & {[[W_{d,2,2}]]_{L}} & \cdots & \vdots \\
		\vdots  & \vdots & \ddots & \vdots \\
		{[[W_{d,R,1}]]_{L}} & \cdots & \cdots & {[[w_{d,R,R}]]_{L}} \\
	\end{array}\right]
	\left[\begin{array}{cccc}
		Q_{1}^{*} & 0_{8\times 1} & \cdots & 0_{8\times 1} \\
		0_{8\times 1} & Q_{2}^{*} & \cdots & 0_{8\times 1} \\
		\vdots  & \vdots & \ddots & \vdots \\
		0_{8\times 1} & \cdots & 0_{8\times 1} & Q_{R}^{*} \\
	\end{array}\right]
	\endgroup
	.
	\end{tiny}
\label{equ:dqmulattentionWithWeights4}
	\end{equation}
Something comparable can be done with equation \eqref{equ:dqmulattentionWithWeights2} through the usage of the alternative matrix representation $[[Q_{d}]]_{R}$ which permutes the multiplication order. Through this operation, the weights move from the outside to the middle of the multiplication chain, allowing them to be combined. 

This procedure excludes weight sharing because it would result in a similarity calculation according to the one proposed in equation \eqref{equ:dqmulattentionKeyQuery}.

\subsubsection{Mapping from Dual Quaternions to Scalar Values}
\label{subsec:MappingDqScalar}

The previous section described how to obtain the attention scores in the dual quaternion space. However, these results can not directly be used since a rating between zero and one is needed as a measurement for the relative importance of an individual input. Therefore, a mapping from the dual quaternion space to a scalar value is needed. For this, several strategies are conceivable, some of them are presented in the following.

\paragraph{Cosine Similarity}

To use the cosine similarity, a target dual quaternion $P_{d, target}$ has to be defined. Then, the similarity between this dual quaternion and one individual attention dual quaternion $A_{d, i}$ can be calculated with
\begin{equation}
	a_{i} = \cos(\phi) = \frac{P_{target} \cdot A_{i}}{\left\| P_{target} \right\| \left\|A_{i}\right\|}
\end{equation}
where both, $P_{d, target}$ and $A_{i}$, are interpreted as an eight-dimensional vector. The smaller the angle $\phi$ between these two vectors in the hyper-space, the more the cosine-similarity approaches the value one. Using a $Softmax()$ on all the obtained attention scores $a_{i}$ afterwards is possible.

\paragraph{Mean-square-error}
Also with this variant, a target dual quaternion $P_{d, target}$ is needed, but this time the mean-square error for the difference between this dual quaternion and one attention dual quaternion $A_{d,i}$, again in vector representation, is used:
\begin{equation}
	d_{i} = (P_{target} - A_{i} )^{T} \cdot (P_{target} - A_{i}) .
\end{equation}
Since $d_{i} \rightarrow 0$ with increasing similarity, this result has the inverse meaning in comparison to the previous one and lies in the range of $[0, \infty]$. Hence, $d_{i}$ has to be brought in the range of $[0, 1]$.
This can be done with 
\begin{equation}
 	a_{i} =	2 \left(\frac{1}{1 + exp(\frac{1}{-d_{i}})} - 0.5\right).
\end{equation}

\paragraph{Error Dual Quaternion}

Another option, but this time completely in dual quaternion space, is to calculate an error dual quaternion $E_{d}$ to a target dual quaternion $T_{d}$ after equation \eqref{equ:errorDQ}:
\begin{equation}
	E_{d,i} = T_{d}^{*}A_{d,i} 
\end{equation}
The real part represents the rotation needed to align the two dual quaternions, the dual part the incorporated translation which can be seen as rotational error and translational error. From the rotation quaternion, only the scalar part is used, which corresponds to $\cos(\frac{\phi}{2})$. The smaller the error angle $\phi$, the more this value reaches one, which is the optimal case. For the translation, the norm of the x-, y- and z-components is used. These two values can be combined to one scalar error value with
\begin{equation}
	a_{i} = \alpha e_{0, i} -  \left\| 2 E^{*}_{i} E_{\epsilon,i} \right\|
\end{equation}
where $\alpha$ is a weighting factor to balance the importance of the angle of rotations $\cos(\frac{\phi}{2})$ contained in $e_{0}$ and the absolute distance from the translation. 

These $a_{i}$ can take values in the range $[-\infty, 1]$, therefore they need to be brought to the range $[0, 1]$ with e.g. the $Sigmoid()$ or using the $Softmax()$ on all $a_{i}$.

\subsubsection{Mapping from Scalar Values to Dual Quaternions}

To multiply the individual attention scores with the input dual quaternions, they need to be brought back to the dual quaternion space. 
This mapping from a scalar value to a dual quaternion can be applied in a very simple way:
one single attention score $a_{d,i}$ is converted with
\begin{equation}
	\Lambda_{d, i} = a_{i}(1 + 0i +0j + 0k + \epsilon0)
\end{equation}
Out of this individual dual quaternions, the whole dual quaternion attention vector $\Lambda_{d}^{g}$ can be formed by concatenating the $\Lambda_{d, i}$ with $\Lambda_{d}^{g} = [\Lambda_{d, 1} \Lambda_{d, 2} \cdots \Lambda_{d, R}]$.\\
\\
In case the attention scores have a matrix form this results in a dual quaternion matrix 
\begin{equation}
	\mathbf{\Lambda}_{d}^{g} = 
	\left[\begin{array}{cccc}
		[[\Lambda_{d,1,1}]]_{L} & [[\Lambda_{d,1,2}]]_{L} & \cdots & [[\Lambda_{d,1,R}]]_{L} \\
		{[[\Lambda_{d,2,1}]]_{L}} & [[\Lambda_{d,2,2}]]_{L} & \cdots & \vdots \\
		\vdots                & \vdots              & \ddots & \vdots \\
		{[[\Lambda_{d,R,1}]]_{L}} & \cdots & \cdots & [[\Lambda_{d,R,2}]]_{L} \\
	\end{array}\right] .
\end{equation}

\subsubsection{Applying the Obtained Attention Scores}
The last required step is to combine the values respectively the inputs with this $\Lambda_{d}^{g}$ or $\mathbf{\Lambda}_{d}^{g}$.
For the vector case, this is done with
\begin{equation}
	C_{d}^{g} = \Lambda_{d}^{g} \circ V_{d}^{g}.
\end{equation}
The straight forward matrix-notation of this is 
\begin{equation}
	C_{d}^{g} = 
	\left[\begin{array}{cccc}
		[[\Lambda_{d,1}]]_{L} & 0_{8\times 8} & \cdots & 0_{8\times 8} \\
		0_{8\times 8} & [[\Lambda_{d,2}]]_{L} & \cdots & 0_{8\times 8} \\
		\vdots                & \vdots              & \ddots & \vdots \\
		0_{8\times 8} & 0_{8\times 8} & \cdots & [[\Lambda_{d,R}]]_{L} \\
	\end{array}\right]
	\left[\begin{array}{cc}
		X_{1} \\ X_{2} \\ \vdots \\ X_{R}
	\end{array}\right]
\end{equation}
When the generated attention result is a matrix, it can be applied directly with
\begin{equation}
	C_{d}^{g} = 
	\mathbf{\Lambda}_{d}^{g} V_{d}^{g}
\end{equation}
or according to the model of the transformer with an additional weight for the values $W_{d}^{g^V}$ by calculating
\begin{equation}
	C_{d}^{g} = 
	\mathbf{\Lambda}_{d}^{g} W_{d}^{g^V} \circ V_{d}^{g}.
\end{equation}
The corresponding matrix notation of the transformer is 
\begin{equation}
\begin{small}
	C_{d}^{g} = 
	\begingroup
	\setlength\arraycolsep{2pt}
	\left[\begin{array}{cccc}
		[[\Lambda_{d,1,1}]]_{L} & [[\Lambda_{d,1,2}]]_{L} & \cdots & [[\Lambda_{d,1,R}]]_{L} \\
		{[[\Lambda_{d,2,1}]]_{L}} & [[\Lambda_{d,2,2}]]_{L} & \cdots & \vdots \\
		\vdots                & \vdots              & \ddots & \vdots \\
		{[[\Lambda_{d,R,1}]]_{L}} & \cdots & \cdots & [[\Lambda_{d,R,R}]]_{L} \\
	\end{array}\right]
	\left[\begin{array}{cccc}
		[[W_{d,1}^{V}]]_{L} & 0_{8\times 8} & \cdots & 0_{8\times 8} \\
		0_{8\times 8} & [[W_{d,1}^{V}]]_{L} & \cdots & 0_{8\times 8} \\
		\vdots                & \vdots              & \ddots & \vdots \\
		0_{8\times 8} & 0_{8\times 8} & \cdots & [[W_{d,1}^{V}]]_{L} \\
	\end{array}\right]	
	\left[\begin{array}{c}
		[[V_{d,1}]]_{L} \\
		{[[V_{d,2}]]_{L}} \\
		\vdots \\
		{[[V_{d,R}]]_{L}} \\
	\end{array}\right] .
	\endgroup
	\end{small}
\end{equation}

\noindent
A comparable architecture can also be implemented with multiplications only under the aspect of unit preservation, similar to the approach from equation \eqref{equ:dqSingleNeuronProd} for the feed-forward architecture. With this method, the attended values $C_{d}^{g}$ are calculated with
%
%\begin{equation}
%C_{i} = \prod_{i=1}^{R} (\mathbf{\Lambda}_{d}^{g} W_{d}^{g^V})_{i}V_{d,i}.
%\end{equation}
%%
%
%\begin{equation}
%	C_{i} = \prod_{i=1}^{R}
%	( W_{d}^{g^K} \circ K_{d}^{g} \circ Q_{d}^{*, g} \circ  W_{d}^{*, g^Q} )  (W_{d}^{g^V} \circ V_{d}^{g} )
%\end{equation}
%
\begin{equation}
	C_{d}^{g} = \Lambda_{d}^{g} \circ (W_{d}^{g^V} \circ V_{d}^{g}) 
\end{equation}
where $\Lambda_{d}^{g}$ is parametrised as
\begin{equation}
	\Lambda_{d}^{g} = [\Lambda_{d, 1} \Lambda_{d, 2} \cdots \Lambda_{d, R} ]^{T},~
	\Lambda_{d, i} = 
	\begin{cases}
		1 + 0i + 0j + 0k + \epsilon 0 & \text{if } a_{i} > 0.5\\
		(W_{d,i}^{V}V_{d,i})^{-1}              & \text{otherwise}
	\end{cases}.
\label{equ:MappingScalarDq}
\end{equation}
For this, $a_{i}$ can be obtained with any of the proposed mappings from dual quaternions to a scalar value or also with a different mapping. The mapping back to dual quaternions is changed to the method from equation \eqref{equ:MappingScalarDq}.

\section{Dual Quaternion Recurrent Neural Networks with Quaternion Attention to predict Rigid Body Dynamics}
\label{sec:DQRNNtoPredictRBD}

In this section, a regular feed-forward neural network to predict rigid body dynamics with a hard coded recurrence on the base of dual quaternions is described. It utilizes an attention mechanism, also operating in dual quaternion space.

\subsection{General Approach}
\label{subsec:GeneralApproach}

The overall goal of the neural network is to predict the movements of rigid bodies in dual quaternion space. For this, the previously described mathematical and physical preliminaries are utilized and combined with the prediction capabilities of neural networks. 

The recurrence in this network architecture will be a hard coded recurrence where the output of one time step forms the input for the next time step, resulting in a closed loop system.This means that none of the widely used known recurrent architectures like standard RNN's, LSTM's or GRU's are used and the neural network itself can be seen as a regular feed-forward one.
\\
\\
As seen in subsection \ref{subsec:rbtDualQuaternions}, a direct transformation of a rigid body described with the combination of its orientation and position in dual quaternion form, based on the model of point transformation, is not possible. Of course each rigid body could be described in such a way that this approach is usable, for example as a set of points which can directly be transformed, but this would eliminate the advantage of the compact and elegant combination of orientation and location in a dual quaternion. 

\begin{figure}[H]
	\centering
	\includegraphics[width=0.9\linewidth]{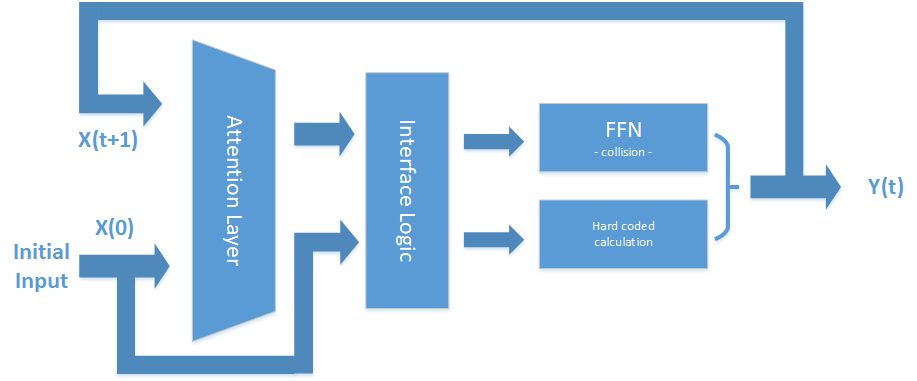}
	\caption{Abstract architecture for the proposed neural network}
	\label{fig:netzSchemaFFN}
\end{figure}

Hence, a different approach where this is possible is chosen. It is based on the kinematic description of rigid bodies and its numerical integration, proposed in the equations \eqref{equ:DqDiffEquation} and \eqref{equ:DqIntegration} in subsection \ref{subsec:RbKinematicsUsingDq}.
\\
\\
For the prediction task we use a two-staged approach in combination with a hard-coded recurrence as shown in figure \ref{fig:netzSchemaFFN}. The attention layer and the following FFN are decoupled using a certain interface logic with the intention to allow a second, alternative path to the FFN and to process the attention output with non-differentiable operations.
To further explain this approach we first need to consider the inputs to the neural network, which will additionally be described in more detail in the following subsection \ref{subsec:dqInputs}. These are namely the position of the rigid body, its orientation, its velocity and angular velocity as well as the momentum and angular momentum, the rigid body's dimensions and finally six planes describing the cuboid simulation environment. All of these inputs share one similarity: a typical rating of soft-attention somewhere in the range of zero to one would falsify their meaning, e.g. a velocity would seem to be lower than it actually is or a plane would be displaced and hence carries no valid information anymore. Therefore, hard-attention could be the way to go, however this means backpropagation can't be used anymore. To overcome this disadvantage, one crucial advantage of the type of data can be used: since the properties of the rigid body in a following time-step are known, it can be extracted if a collision with a bounding wall will occur, and if so where it will occur. This allows the definition of targets for the attention layer and hence separate training and processing of the attention output to scores of either one or zero without hurting the overall differentiability. \\
\\
For the prediction the neural network has to do, two general cases are possible, which directly affect the desired attention output: either the movement of the rigid body under consideration is affected by a collision or it is not. For the first case, this means all inputs describing the rigid body and furthermore one bounding wall description are relevant for the prediction and need to be rated with a one, the other planes need to be rated with a zero. The second case is way easier, there all planes describing the simulation area are irrelevant, just like the dimensions of the rigid body. These desired attention outputs can be rated as an indirect collision detection, because the rating of the bounding walls determines which of the two possible cases applies for the prediction of the next time step.

This intermediate result enables the incorporation of another, strong inductive bias beside just the simple usage of dual quaternions: according to equation \eqref{equ:DqTwist}, the behaviour of the rigid body is already known, hence it can directly be calculated with a second hard coded path, a prediction with the FFN is only necessary in the case of a collision.

\subsection{Inputs and their Dual Quaternion Representation}
\label{subsec:dqInputs}

For the proposed prediction approach, a total number of  13 inputs shall be used. These are the already named position of one rigid body's centre of mass, its orientation, its velocity and angular velocity, the rigid body's dimensions as length, width and height, its impulse, its angular momentum and the six walls of the bounding area. 

Contrary to the integration approach with the pose, thus the combination of position and orientation, both are separate inputs to eliminate the work of separating them for the network, simplifying the overall task a little bit.\\
\\
The position of the centre of mass is encoded as
\begin{equation}
	P_{d} = 1 + \epsilon(p_{x}i + p_{y}j + p_{z}k)
\end{equation}
and the rigid body's orientation as
\begin{equation}
	O_{d} = o_{0} + o_{1}i + o_{2}j + o_{3}k
\end{equation}
where $o_{0}, o_{1}, o_{2}, o_{3}$ are the four parts of a regular unit quaternion representing orientations and the dual part is zero.
The next inputs are the rigid body's velocity
\begin{equation}
	V_{d} = 0 + \epsilon (v_{x}i + v_{y}j + v_{z}k)
\end{equation}
causing the translation of a rigid body, and the angular velocity
\begin{equation}
W_{d} = w_{x}i + w_{y}j + w_{z}k
\end{equation}
causing the rotation of the rigid body. 

Also required are the dimensions
\begin{equation}
	G_{d} = g_{1}i + g_{2}j + g_{3}k
\end{equation}
where $g_{1}$ corresponds to half of the length,  $g_{2}$ to  half of the width, and  $g_{3}$ to half of the height of the rigid body, because the centre of mass alone is not sufficient for collision detection and predicting its impact in any form. 

The impulse $I_{d}$ and angular momentum $L_{d}$ of a rigid body are denoted just like their velocity counterpart as 
\begin{equation}
I_{d} = 0 + \epsilon (i_{x}i + i_{y}j + i_{z}k)
\end{equation}
and 
\begin{equation}
L_{d} = l_{x}i + l_{y}j + l_{z}k.
\end{equation}
The last missing inputs are the six bounding walls $\mathcal{S}_{d}$, which are described as a plane $S_{d}$ in dual quaternion space. For a single wall, two components described as pure quaternions are needed: an arbitrary point $\textbf{z} = z_{x}i + z_{y}j + z_{z}k$ located on the plane and a unit norm vector $\textbf{n} = n_{x}i + n_{y}j + n_{z}k$ which form the wall description \cite{Radavelli.2014}
\begin{equation}
	S_{d}^{(m)} = \textbf{n}^{(m)} + \epsilon( \textbf{z}^{(m)} \cdot \textbf{n}^{(m)}), ~\forall~m \in \mathcal{S}_{d} .
\end{equation}

\subsection{Attention stage}

As already stated in subsection \ref{subsec:GeneralApproach}, the desired attention output only differs regarding the bound walls and the description of the rigid body's dimensions. The later one is furthermore indirectly determined by the attention rating of the bounding walls, it is only irrelevant and to be rated with a zero if all walls are rated with zero in the case of no collision.
Hence, the prediction task can be simplified to only target the bounding walls and omit the other inputs. To take up the idea of indirect collision detection again, this prediction task can be also seen from a classification problem point of view.

Therefore, we propose two different versions for the attention stage, one based on a multi-classifier approach and a second one based on a binary-classifier approach. Section \ref{sec:Experiments} will include an experimental comparison of both approaches.

\subsubsection{Multi-Classifier}

In the multi-classifier approach, each bounding wall gets assigned a imaginary class, additionally the case "no collision" becomes a further seventh class. The neural network is a feed-forward neural network as described in equation \eqref{equ:dqnnFfnGeneral} with two hidden layer with biases and an element-wise operating activation function. The best working activation is also experimentally determined in section \ref{sec:Experiments}. To prevent potential overfitting, a dropout layer is incorporated before each of the two hidden layers. The output layer consists of seven dual quaternion neurons, each representing one of the classes. \\
\\
The desired output is the probability of each class being the one fitting to the target. Therefore, the dual quaternion has to be converted to a scalar since it cannot represent a probability itself as described in subsubsection \ref{sec:DqAttention}. For this, the one-equivalent dual quaternion $1 + 0i + 0j + 0k + \epsilon(0 + 0i + 0j + 0k)$ is defined as maximum probability, and a similarity score to this dual quaternion forms the predicted probability. This score is calculated in the following way: initially, the seven output dual quaternions are normalized with equation  \eqref{equ:dqNormalization}, subsequentially with this result an error dual quaternion as proposed in subsection \ref{subsec:MappingDqScalar} is calculated. For this, an $\alpha$ of 100 is used and the values are normalized with a $Softmax()$.

Since the output represents a probability, no mapping back to the dual quaternion space is necessary. During training, the cross-entropy loss is used to optimize the networks parameter. The propability result is later processed with the $max()$ operator which selects the class with the highest predicted probability, resulting in the wall which is attended or the information that there is no collision with a wall to expect.

\subsubsection{Binary-Classifier}
\label{sec:BinaryClassifier}

Instead of targeting all possible collisions at once, this can be done for each possibility separately, too.  This means, that for all six walls, one individual run through a shared neural network is done with only this specific wall as part of the input values. The desired output is a binary value, answering the question "is a collision with this specific wall to be expected?".
This allows for a massive weight sharing and eventually a smaller net in comparison to the multi-classifier approach, making the model less susceptible to potential overfitting. 

This network also uses two hidden layer with biases and dropout in front to prevent potential overfitting. Likewise to the multi-classifier approach, the activation functions operate element-wise and the best working one is determined later in section \ref{sec:Experiments}.

The shared net only has one output neuron, where an output of zero is targeted for no collision and an output of one shall indicate a collision. To map the dual quaternion output to a scalar value, the same strategy as for the multi-classifier is used. Because of the single output, this time the binary-cross-entropy loss is chosen as the loss function. \\
\\
Nevertheless, this six successively calculated outputs don't represent the case of no collision to be expected. To make up for this, a threshold of 0.4 is set as a minimum requirement for an output to be rated as one. When all predicted probability values are relatively low and stay under this desired threshold, this indicates that no collision with a bounding wall is to be expected.

\subsection{Collision stage}
As already stated, the sequence-generation approach is based on the kinematic description of rigid bodies and hence on the twist and its numerical integration.

Therefore, the collision stage is designed as a dual quaternion feed forward neural network predicting the twist as a dual quaternion for the case of a collision of the rigid body under consideration with any obstacle. 

\begin{figure}[H]
	\centering
	\includegraphics[width=0.5\linewidth]{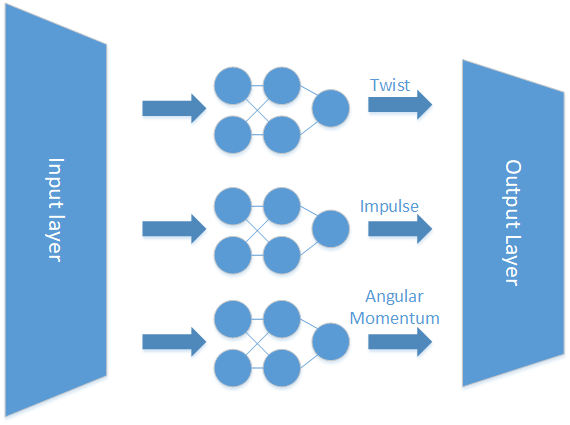}
	\caption{Abstract architecture for the proposed neural network}
	\label{fig:netzSchemaCollisionStage}
\end{figure}

Two characteristic quantities that describe the state of a rigid body are the velocity and angular velocity, and usually they are expected to change during a collision. They can be calculated with the rigid body's impulse $I_{d}$ and angular momentum $L_{d}$. Unfortunately both these quantities can't be extracted from the rigid body's pose, making two more predicted dual quaternions required. For this, two additional and independent paths inside the network are used since twist, impulse and angular momentum are not directly physically related to each other, therefore they might need different intermediate results in the hidden layers. These three paths are in fact separate feed-forward neural networks and only share the common inputs and the output layer. This leads to the overall structure for the collision stage shown in image \ref{fig:netzSchemaCollisionStage}. 

\subsubsection{Output Layer}
\label{sec:outputLayer}

The output layer is not a classical neural network layer containing neurons, in fact there aren't even trainable parameters existent in this layer. Instead, it has the function to do the necessary calculations to map the predictions to the new input for the following time step and an updated rigid body description. 
%
%The predicted twist $\hat{\xi}$ is calculated with the neural networks weights in dual quaternion space $W$ and the input to the network $X$, containing $P_{d}, V_{d}, W_{d}, G_{d}$ and $\mathcal{U}_{d}$
%\begin{equation}
%	\hat{\xi}_{d} = WX
%\end{equation}
%%
\\
\\
With the predicted twist $\hat{\xi}_{d}$ and equation \eqref{equ:DqIntegration} the new pose $	P_{d}(t+\Delta t)$ of the rigid body is calculated with:
\begin{equation}
	P_{d}(t+\Delta t) = exp(\hat{\xi}_{d} \dfrac{\Delta t}{2})P_{d}(t).
\label{equ:poseIntegrationStep}
\end{equation}
\\
To do so, prior to this, position and orientation have to be combined to the pose and afterwards split again. This can be done with either equations \eqref{equ:dqRotTrans} and \eqref{equ:extractTrans} or with equations \eqref{equ:dqTransRot} and \eqref{equ:extractTrans2}.

For the predictions of the next time step, also the velocities $V_{d}$ and $W_{d}$ need to be updated since they are generally not constant over time, making additional calculations necessary.
\\ 
\\
The calculation of $V_{d}(t+\Delta t)$ is straight forward with
%%
%\begin{equation}
%	V_{d}(t+\Delta t) = \frac{T(t+\Delta t) - T(t)}{2}
%\end{equation}
%%
\begin{equation}
	V_{d}(t+\Delta t) = \frac{\hat{I}_{d}}{m_{rigid~body}}.
\end{equation}
The new angular velocity can't be calculated comparably straight forward. Instead, the physical properties of the rigid body kinematics, described in subsection \ref{subsec:rigidBodyKinematics}, have to be used.

As already known, the relationship of angular momentum $L$ and angular velocity is
\begin{equation}
	L = \textbf{I}_{world}\cdot \omega
\label{equ:relAngMomAngVel}
\end{equation}
and $\textbf{I}_{world}$ can be calculated with
\begin{equation}
	\textbf{I}_{world} = R\cdot \textbf{I}_{body}\cdot R^{T}
\end{equation}
where $R$ is a rotation matrix obtained from the real part $O$ of the orientation dual quaternion $O_{d}(t+\Delta t)$ describing the attitude of the rigid body.

An alternative way of calculating $\textbf{I}_{world}$ directly with the attitude quaternion $O$ is \cite{Zhao.2013}
\begin{equation}
	\textbf{I}_{world} = (O(OI_{body}O^{*})^{T}O^{*})^{T}
\end{equation}
where 
\begin{equation}
	\textbf{I}_{body} = 
	\left[\begin{array}{cccc}
		1 & 0 & 0 & 0 \\
		0 & I_{1} & 0 & 0  \\
		0 & 0 & I_{2} & 0 \\	
		0 & 0 & 0 & I_{3} \\
	\end{array}\right]
\end{equation}
and each column of the I-Matrix is treated as a quaternion. This calculation can be interpreted as first rotating all the columns and afterwards all the rows of I to obtain the matrix rotated with a quaternion.

These rotation can be extended to dual quaternions by choosing
\begin{equation}
	\textbf{I}_{body} = 
	\left[\begin{array}{cccccccc}
		1 & 0 & 0 & 0 & 0 & 0 & 0 & 0 \\
		0 & I_{1} & 0 & 0 & 0 & 0 & 0 & 0 \\
		0 & 0 & I_{2} & 0 & 0 & 0 & 0 & 0 \\	
		0 & 0 & 0 & I_{3} & 0 & 0 & 0 & 0 \\
		0 & 0 & 0 & 0 & 1 & 0 & 0 & 0 \\
		0 & 0 & 0 & 0 & 0 & 1 & 0 & 0 \\
		0 & 0 & 0 & 0 & 0 & 0 & 1 & 0 \\
		0 & 0 & 0 & 0 & 0 & 0 & 0 & 1 \\
	\end{array}\right]
\end{equation}
and using the orientation dual quaternion $O_{d} = O_{d}(t+\Delta t)$ such that
\begin{equation}
\textbf{I}_{world} = (O_{d}(O_{d}I_{body}O_{d}^{*})^{T}O_{d}^{*})^{T} .
\end{equation}
Hence, the new angular velocity $\omega(t+\Delta t)$ can be obtained with
\begin{equation}
	\omega(t+\Delta t) = \textbf{I}_{world}^{-1}(t+\Delta t)\cdot \hat{L} 
\end{equation} 
%
%In the case that the angular momentum $L$ is not known, it can be calculated using equation \eqref{equ:relAngMomAngVel}, the angular velocity $\omega(t)$ and the moment of inertia tensor $\textbf{I}_{world}(t)$ in world coordinates. This leads to the relation
%%
%\begin{equation}
%\omega(t+\Delta t) = \textbf{I}_{world}^{-1}(t+\Delta t)\cdot \textbf{I}_{world}(t)\omega(t) 
%\end{equation} 
%%
%\\
%\\
%Since $G_{d}$ and $\textbf{I}_{body}$ are constant they don't need to be updated.
The parameters $m_{rigid~body}$ and $\textbf{I}_{body}$ are constants and part of the dataset, but since they are not needed for the prediction they are not included in the input set. Instead, they are additional parameters passed to the neural network just once at the beginning.

\subsection{Interface Layer}

In the interface layer, it has to be separated between the two proposed classification approaches in the attention stage: multi- or binary-classification. In multi-classification, the approach is straight forward: the output corresponds directly to a probability distribution which event is most likely to happen, therefore choosing the highest value with a $max()$ operator is the likeliest event.

For binary-classification the six attention-output values are independent from each other, they don't sum up to one. Hence it can happen that in case of a bad prediction, two or even more high values, meaning a high likelihood, can be a result. Anyway, an unambiguous result is needed here since it is not intended to pass multiple walls to the prediction in the collision stage. Therefore, equally a $max()$ operation is used. Furthermore, there is the possibility for no collision, which is not directly present in a high probability coming from the six output neurons. Instead, this is the case when all predicted probabilities are low. Therefore, an additional threshold is incorporated and has to be exceeded before the $max()$ operation is done. If this is not the case, no collision is expected. 

If the attention predicts a collision, the input set is split into everything but the wall dual quaternions and solely the wall dual quaternions. Then, the relevant wall is selected and concatenated with the remaining dual quaternions to form the input set for the prediction task. Thus, the number of inputs  reduces by five in comparison to the input to the attention stage. In case that there is no collision detected, the inputs are instead passed to the hard coded function to calculate the twist and integration step, the prediction with the neural network is skipped.

\subsection{Training Procedure}

Both networks are trained separately, using an independent loss function especially suitable for their outputs. After training, the learnable parameters producing the best results in a validation dataset are saved so that they can later be loaded in the combination of both networks.

Since the number of inputs is reduced by the attention and interface layer, the same has to be done for the training dataset for the collision stage. To avoid handling two datasets, the labels used for training the attention network can furthermore be utilized to act as a perfectly working attention mechanism, such that the same logic from the interface layer can be used to achieve this reduction. All data points where no collision takes place are simply skipped for the training of this network when loading the data.
\\
\\
An end-to-end training is not intended for the time being since it would limit the logic in the interface layer to differentiable operations which is not possible with the proposed concept of hard-removing e.g. the five walls and using $max(\cdot)$ operations. Furthermore the two hard-switched paths would cause further problems in differentiation.

\section{Experiments}
\label{sec:Experiments}

As an initial proof of concept, we consider a very basic simulation setup with a fixed rigid body dimension and also a fixed simulation environment, even though we are aware of the very limited generalization capabilities. With the proposed architecture, different hyperparameter-configurations are evaluated.

\subsection{Dataset}

The used dataset was generated with the DEM-simulation LIGGGHTS from the DFDEM\textregistered project \cite{CFDEM} and the included superquadric particle shape. In this simulation the rigid body dynamics are calculated based on initial conditions in a given simulation environment. It uses discrete time steps of $1\times 10^{-5}s$, and every $n$ time steps an output describing the simulated rigid bodies can be issued 

For this, every $100ms$ an output file is issued, containing the following information to describe a rigid body: a quaternion describing the orientation, the position of the  centre of mass in x-, y- and z-coordinates, the velocity and angular velocity in x-, y- and z-components, 
the dimensions as half of the length, width and height, the mass and the angular momentum in x-, y- and z-components. In total, 150000 items were included.
\\
\\
Since this is a completely novel approach and neural network architecture, the complex task of predicting rigid body dynamics shall be simplified as much as possible for the first experiments. To do so, only one cube-shaped rigid body with rounded corners and the length, with and height of $0.2m$ is used. 

The simulation environment is also chosen as a cube with a length, width and height of $0.4m$ where the origin of the global coordinate system is located in the centre of the cube.   The environment is deliberately chosen so small in relation to the body to enlarge the amount of contacts with the walls in relation to the free movement. This is necessary because of the small time steps, needed for integration accuracy, resulting in many data points for the free movement. Furthermore, the cube-shape favours the augmentation which is introduced in the following section.
 
%To determine the moment of inertia tensor in body coordinates, separate simulation runs are done where the rigid body is successively rotated around its principal axis of inertia. This is necessary because the round edges of the rigid body complicate the geometric calculation.

\subsection{Data Augmentation}

Machine learning tasks are highly dependent on big training datasets to achieve high performance and a good generalization capability. These are often not existent, resulting in worse training results and sometimes overfitting models.
Data augmentation targets this by artificially expanding the dataset through transformations on the original data. This can be done with various techniques: very popular in image classification tasks is e.g. flipping or rotating the image, manipulating the images' color space, cropping the image or randomly erasing parts of the image. Also further advanced techniques like mixing images or generating additional samples using GANs are possible and used. \cite{Shorten.2019}   

Augmentation is also used in other machine learning areas like speech recognition. Here, e.g. selected consecutive frequencies or consecutive time steps are masked \cite{Park.}. 
\\
\\
In this work, the idea of geometric augmentations from image classification tasks is transferred to the three dimensional space. Furthermore, the ability of easy rotations with quaternions or dual quaternions is utilized. 

Say a collision of a rigid body occurs with the bounding wall in the positive x-direction of the global coordinate system. Then the collision would take place in the same manner if the rigid body and all its properties would be rotated for e.g. $180^{\circ}$ around the x-axis, just upside-down. The same mind-game can also be done for rotating $\pm90^{\circ}$  around the z-axis, what would bring the collision to the wall in the positive respectively negative y-axis direction, and multiple other rotations. The whole process can be also seen as rotating the global coordinate system instead of the rigid body, it only changes the notation but not the collision.

Therefore, to artificially increase the dataset items which belong to a collision, this rotation strategy is used. Since all values describing a rigid body are dual quaternions, after defining the rotation dual quaternion this is done with a simple calculation as introduced in subsection \ref{subsec:rbtDualQuaternions} with a translation $T = 0$. 
Specifically eight augmentations or rotations are done: 3 rotations of $90^{\circ}, 180^{\circ}$ and $270^{\circ}$ around the axis pointing to the wall where the "original" collision takes place and five more to project the collision to the five other walls of the bounding cube. Now it is also clear why the simulation environment is chosen as a cube with the origin of the global coordinate system in its centre. For this reason the proposed augmentation is always possible, otherwise complicated and potentially error-prone translations would have to be incorporated to make up for a non cubic-shape.

\subsection{Experiment execution and results}
\label{subsec:ExperimentResults}
Since the proposed architecture is composed out of two separate trainable neural networks, they are initially threated independently from each other. Only in the following prediction evaluation in \ref{subsec:predictionEvaluation} they are combined with the interface layer and the hard coded recurrence. The following experiments itself are only based on one-step predictions.

\subsubsection{Attention stage}
For the attention stage, both proposed architectures are used and tested against each other. To do so, a two-step procedure was used. Out of an initial search space, the most promising configurations were identified and used to define a second, more precise search space to determine the final neural network model. For the attention stage, the initial search space was the following:

\begin{itemize}
    \item $TanH$, $TanShrink$ and $ReLU$ as activation functions
    \item dual quaternion neuron numbers per layer between 8 and 64
    \item learning rates of $1\times 10^{-2}$ and $1\times 10^{-3}$
    \item dropout rates of 0, 0.1 and 0.2
\end{itemize}

\begin{figure}[H]
	\begin{subfigure}{0.5\textwidth}
		\centering
		\includegraphics[width=1.0\linewidth]{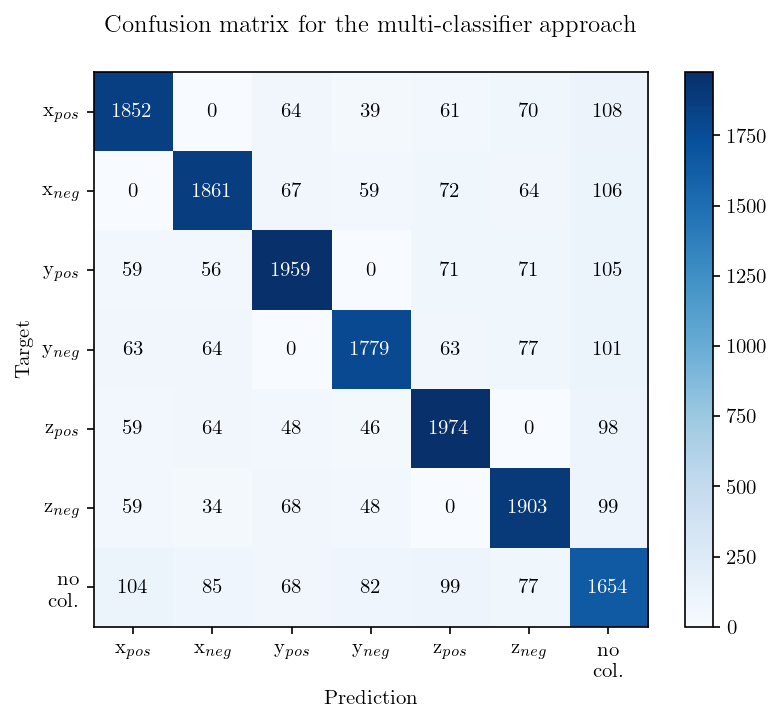}
		\caption{multi-classifier approach}
		\label{subfig:confMatrices_secondGridSearch_multi}
	\end{subfigure}
	\begin{subfigure}{0.5\textwidth}
		\centering
		\includegraphics[width=1.0\linewidth]{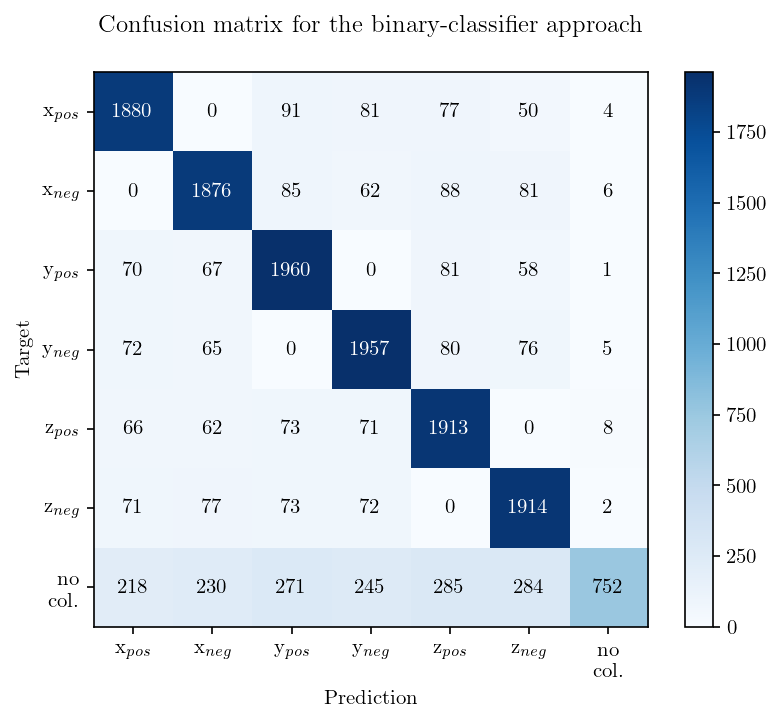}
		\caption{binary-classifier approach}
		\label{subfig:confMatrices_secondGridSearch_binary}
	\end{subfigure}
	\caption{Confusion matrices for the best attention networks predicting whether there is a collision with one of the six bounding walls and if so where}
	\label{fig:confMatrices_secondGridSearch}
\end{figure}

For the second search, also learning rate schedulers were introduced, namely a exponential learing rate decay of 0.9995 after each epoch and a step-wise halving of the learning rate after 1000 epochs each. Also higher dual quaternion neuron numbers of 80 and 96 were introduced for the multi-classifier approach, furthermore only the $TanH$ was used as an activation function as this showed the best results in the first step. This led to the five best results for shown in tables \ref{tab:secondGridsearchAttentionMulti}  and \ref{tab:secondGridsearchAttentionBinary}.
\begin{table}[h!]
	\caption{The five best results for the attention network based on the multi-classifier approach}
    \vspace{8pt}
	\begin{center}
		\begin{tabular}{c c c c c c c}
			\hline
			neurons  & neurons  & dropout & learning & scheduler &  training & testing \\
			layer 1  &  layer 2 & rate    & rate     &           &  accuracy & accuracy \\
			\hline
			96 & 96 & 0.2 & 0.01 & ExponentialLR & 87.02\% & 83.69\% \\
			96 & 80 & 0.1 & 0.01 & ExponentialLR & 91.10\% & 83.64\% \\
			96 & 80 & 0.1 & 0.01 & StepLR        & 91.67\% & 83.48\% \\
			96 & 64 & 0.1 & 0.01 & StepLR        & 90.78\% & 83.39\% \\
			64 & 64 & 0.1 & 0.01 & StepLR        & 89.11\% & 83.33\% \\
			\hline
		\end{tabular}
	\end{center}
	\label{tab:secondGridsearchAttentionMulti}
\end{table}
\begin{table}[h!]
	\caption{The five best results for the attention network based on the binary-classifier approach}
	\vspace{8pt}
    \begin{center}
		\begin{tabular}{c c c c c c c}
			\hline
			neurons  & neurons  & dropout & learning & scheduler &  training & testing \\
			layer 1  &  layer 2 & rate    & rate     &           &  accuracy & accuracy \\
			\hline
			32 & 32 & 0.1 & 0.01  & StepLR        & 78.50\% & 78.95\% \\
			32 & 32 & 0.1 & 0.01  & None          & 75.43\% & 78.55\% \\
			24 & 24 & 0.1 & 0.01  & StepLR        & 77.94\% & 78.02\% \\
			24 & 24 & 0.1 & 0.01  & None          & 73.34\% & 77.83\% \\
			32 & 24 & 0.1 & 0.01  & StepLR        & 77.02\% & 77.63\% \\
			\hline
		\end{tabular}
	\end{center}
	\label{tab:secondGridsearchAttentionBinary}
\end{table}
Furthermore, the two confusion matrices in figure \ref{fig:confMatrices_secondGridSearch} show tendencies which predictions are good and for which predictions mistakes are made.
It is striking that, especially in the binary approach, the most errors are made in the prediction of the case "no collision", what could bear problems when both models are combined to predict a sequence of rigid body motion.

\subsubsection{Collision stage}
The same two-step procedure as for the attention stage was used for the collision stage, too, with the exact same set of hyper-parameters for the first step. Again, in the second step the two schedulers were added but contrary to the previous collision stage this time the $TanhShrink$ was used as an activation function. This led to the five best results for each path as shown in tables \ref{tab:nexGridsearchTwist}, \ref{tab:nexGridsearchImp} and \ref{tab:nextGridsearchAngmom}.
\begin{table}[h!]
	\caption{The five best results for the predicted twist}
	\label{tab:nexGridsearchTwist}
	\vspace{8pt}
    \begin{center}
		\begin{tabular}{c c c c c c}
			\hline
			neurons  & neurons  & dropout & learning & scheduler &  validation \\
			layer 1  &  layer 2 & rate    & rate     &    &  loss \\
			\hline
			80 & 80 & 0.1 & 0.001 & StepLR & 0.992 \\
			80 & 64 & 0.1 & 0.001 & StepLR & 0.993 \\
			80 & 60 & 0.1 & 0.001 & ExponentialLR & 0.993 \\
			80 & 64 & 0.1 & 0.001 & ExponentialLR & 0.994 \\
			64 & 64 & 0.1 & 0.001 & StepLR & 0.998 \\
			\hline
		\end{tabular}
	\end{center}
\end{table}
\begin{table}[h!]
	\caption{The five best results for the predicted impulse}
	\label{tab:nexGridsearchImp}
	\vspace{8pt}
    \begin{center}
		\begin{tabular}{c c c c c c}
			\hline
			neurons  & neurons  & dropout & learning & scheduler &  validation \\
			layer 1  &  layer 2 & rate    & rate     &    &  loss \\
			\hline
			64 & 32 & 0.1 & 0.01 & ExponentialLR & 0.0784 \\
			80 & 48 & 0.1 & 0.01 & StepLR & 0.0796 \\
			80 & 64 & 0.2 & 0.01 & ExponentialLR & 0.0797 \\
			64 & 32 & 0.1 & 0.01 & StepLR & 0.0798 \\
			80 & 80 & 0.1 & 0.01 & ExponentialLR & 0.0800 \\
			\hline
		\end{tabular}
	\end{center}	
\end{table}
\begin{table}[h!]
	\caption{The five best results for the predicted angular momentum}
	\label{tab:nextGridsearchAngmom}
	\vspace{8pt}
    \begin{center}
		\begin{tabular}{c c c c c c}
			\hline
			neurons  & neurons  & dropout & learning & scheduler &  validation \\
			layer 1  &  layer 2 & rate    & rate     &           &  loss \\
			\hline
			64 & 64 & 0.1 & 0.001 & StepLR        & 0.551$\times 10^{-3}$ \\
			80 & 80 & 0.1 & 0.001 & StepLR        & 0.551$\times 10^{-3}$ \\
			80 & 64 & 0.1 & 0.001 & StepLR        & 0.552$\times 10^{-3}$ \\
			48 & 32 & 0.1 & 0.001 & ExponentialLR & 0.552$\times 10^{-3}$ \\
			64 & 32 & 0.1 & 0.001 & StepLR        & 0.552$\times 10^{-3}$ \\
			\hline
		\end{tabular}
	\end{center}	
\end{table}

\subsection{Prediction Evaluation}
\label{subsec:predictionEvaluation}
\begin{figure}[h!]
	\centering
	\includegraphics[width=0.95\linewidth]{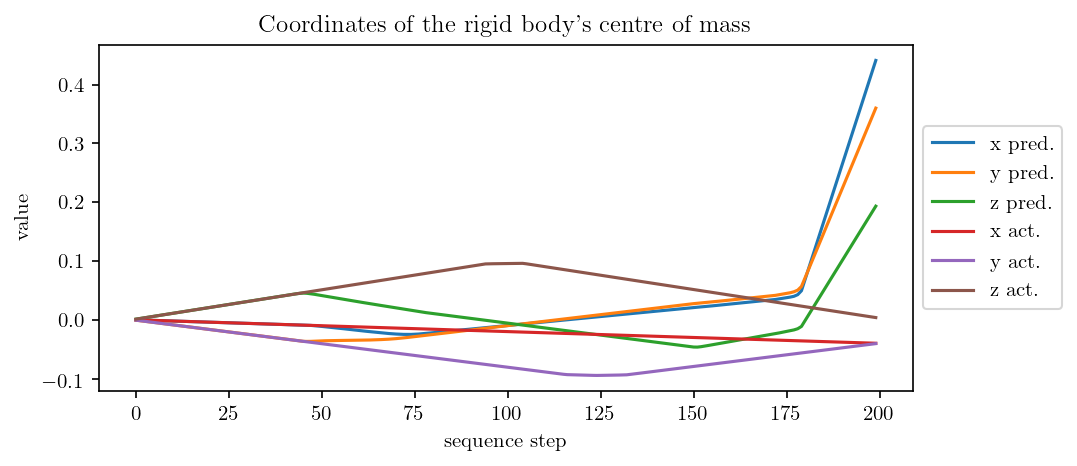}
	\caption{Coordinates of the rigid body's centre of mass calculated by integrating the twist and the actual centre of mass}
	\label{fig:predCoordinatesSeqAtt}
\end{figure}

\begin{figure}[h!]
	\centering
	\includegraphics[width=0.95\linewidth]{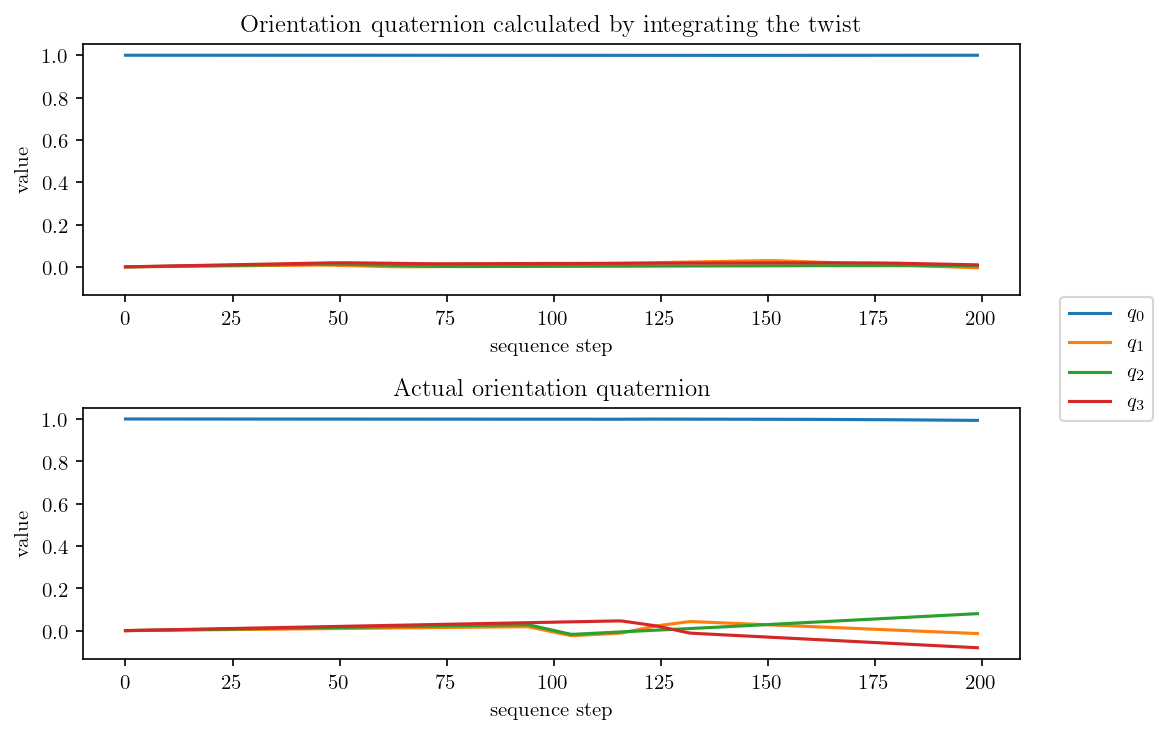}
	\caption{Orientation quaternion calculated by integrating the twist and the actual orientation quaternion}
	\label{fig:predOriSeqAtt}
\end{figure}

\begin{figure}[h!]
	\centering
	\includegraphics[width=0.95\linewidth]{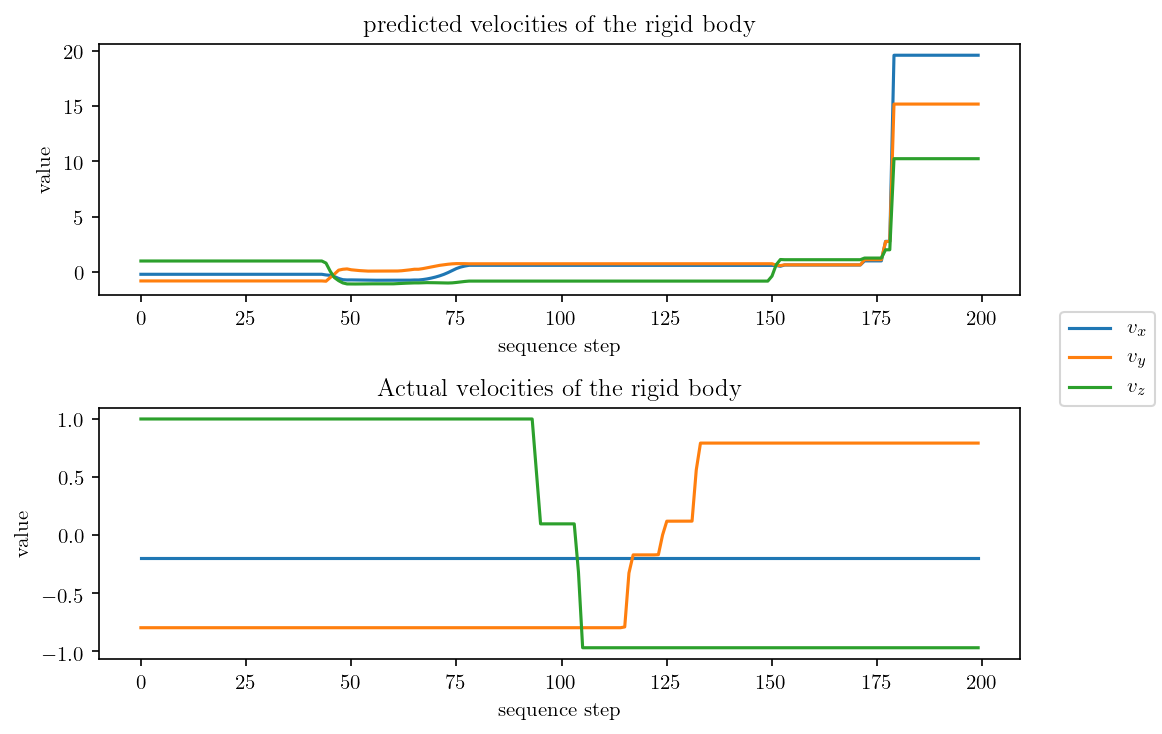}
	\caption{Velocities calculated with the predicted impulse and the actual velocities}
	\label{fig:predVelSeqAtt}
\end{figure}

\begin{figure}[h!]
	\centering
	\includegraphics[width=0.95\linewidth]{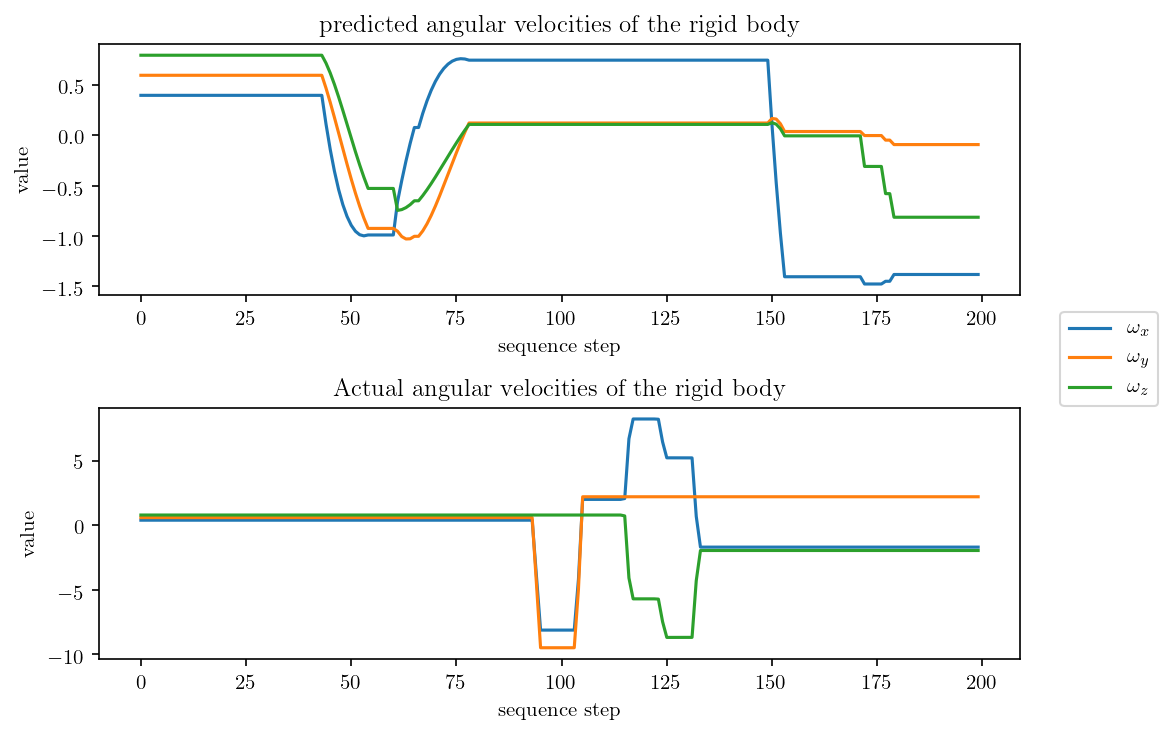}
	\caption{Angular velocities calculated with the predicted angular momentum and the actual angular velocities}
	\label{fig:predAngVelSeqAtt}
\end{figure}

With the previously determined trained models the prediction capabilities can be evaluated. For this, the ability to predict sequences of rigid body motions is considered. The initial conditions are chosen from the first element of the used dataset.The performance measurements are the coordinates of the rigid body's centre of mass, the velocities and angular velocities as well as the quaternion describing the attitude of the rigid body.
Figure \ref{fig:predCoordinatesSeqAtt} shows the trajectory of the centre of mass which results from integrating the predicted twists $\hat{\xi}_{d}$. This integration also directly determines the orientation of the rigid body, which is shown as a single quaternion by just omitting the zero dual part in figure \ref{fig:predOriSeqAtt} 
Additionally to the twist, also the rigid body's impulse and angular momentum is predicted for each time step to determine the corresponding velocity and angular velocity, which are needed as an input for the following prediction. The calculated velocity is shown in figure \ref{fig:predVelSeqAtt} and the angular velocity in figure \ref{fig:predAngVelSeqAtt}. 
Initially, there is no visible difference between the predicted and the actual movement of the rigid body. This is due to the fact that the exact physical relation is used to achieve maximum accuracy. 
However, the attention network mistakenly detected a collision just after 40 time steps and the following collision network caused a change in the trajectory of the rigid body, yielding wrong sequence steps after this point. Small changes in the velocities and angular velocities indicate that every now and then alleged collisions are expected from the attention network are predicted even though the rigid body's centre of mass is in an uncritical area.

Just after 175 time steps of simulation, the rigid body has undergone such a change that it left the simulation environment without a detected collision. After this point, the prediction is done in an domain the network is not trained on so that meaningful results can no longer be expected. The fact that the orientation quaternion in the predicted sequence has undergone almost no changes is explainable with the low angular velocities. Visible changes in the target sequence occur also only after the angular velocities rise to higher values.

\begin{figure}[h!]
	\centering
	\includegraphics[width=0.95\linewidth]{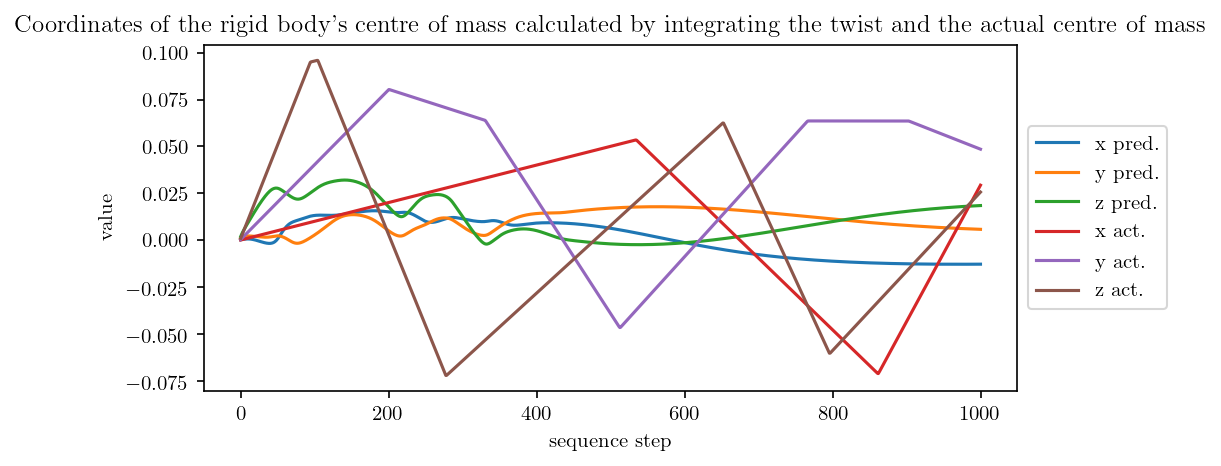}
	\caption{Coordinates of the rigid body's centre of mass calculated by integrating the twist and the actual centre of mass}
	\label{fig:predCoordinatesSeq}
\end{figure}

\begin{figure}[h!]
	\centering
	\includegraphics[width=0.95\linewidth]{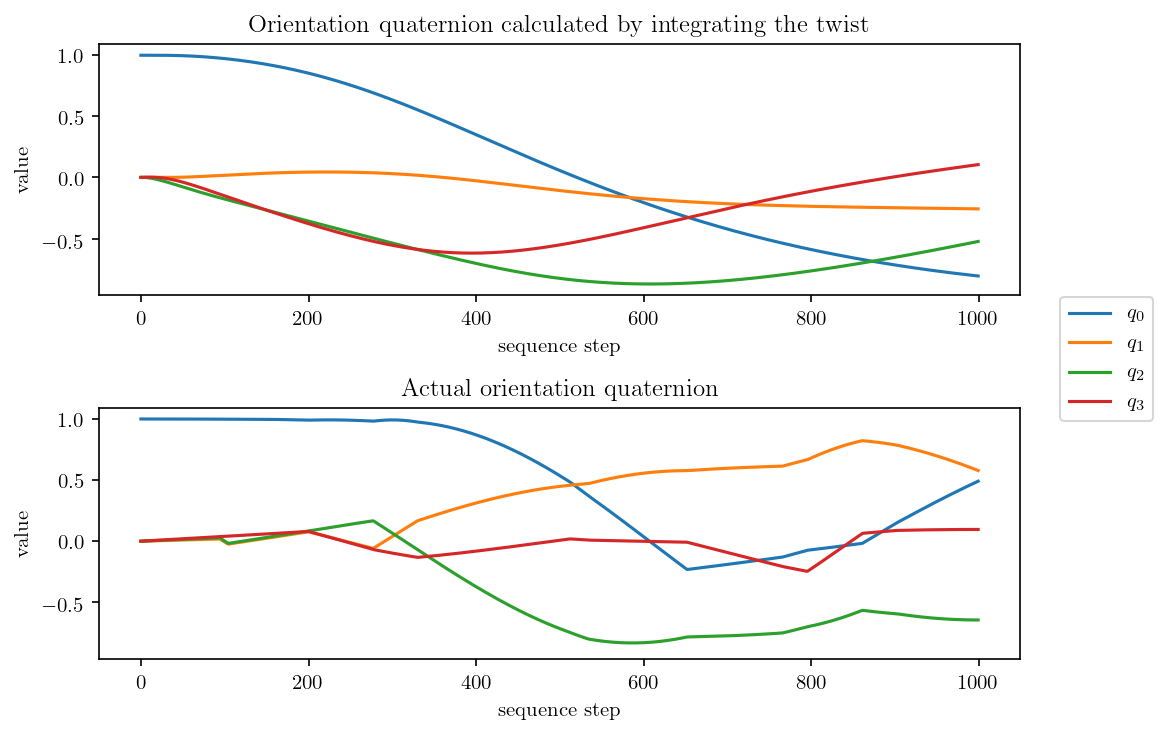}
	\caption{Orientation quaternion calculated by integrating the twist and the actual orientation quaternion}
	\label{fig:predOriSeq}
\end{figure}

\begin{figure}[h!]
	\centering
	\includegraphics[width=0.95\linewidth]{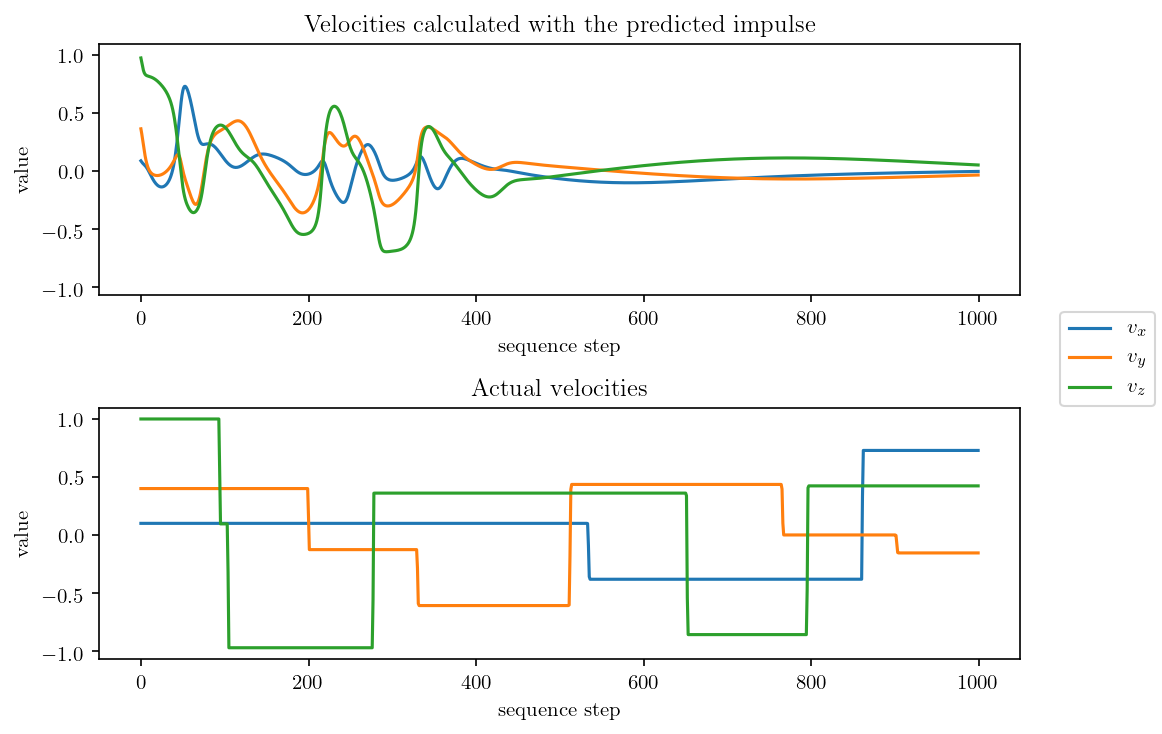}
	\caption{Velocities calculated with the predicted impulse and the actual velocities}
	\label{fig:predVelSeq}
\end{figure}

\begin{figure}[h!]
	\centering
	\includegraphics[width=0.95\linewidth]{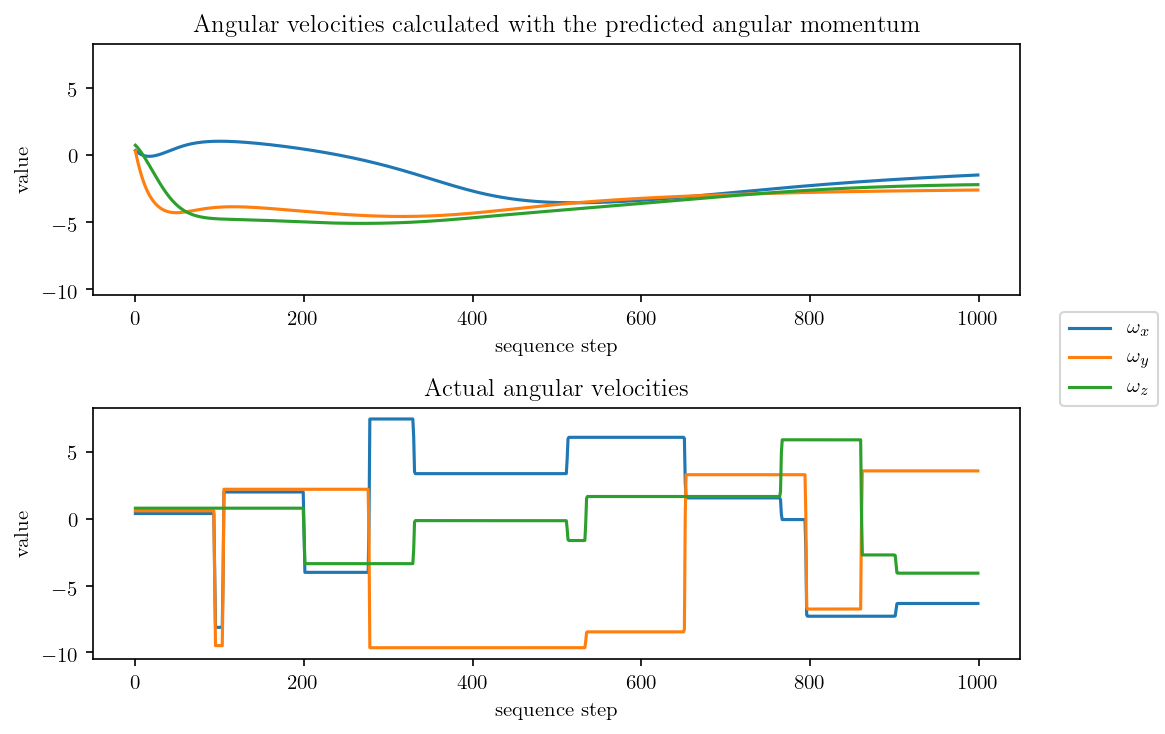}
	\caption{Angular velocities calculated with the predicted angular momentum and the actual angular velocities}
	\label{fig:predAngVelSeq}
\end{figure}

As a comparison, the same prediction was done with a regular FFN without the usage of attention and the second hard coded path, particularly the overall architecture was identical to just the collision stage. A good hyperparameter-configuration was determined with the same two-stage approach as in subsection \ref{subsec:ExperimentResults}. Figure \ref{fig:predCoordinatesSeq} shows the trajectory of the centre of mass, figure \ref{fig:predOriSeq} the orientation quaternion and figures \ref{fig:predVelSeq} and \ref{fig:predAngVelSeq} the course of the calculated velocities and angular velocities.
It can be observed that just after a few prediction steps, the rigid body's centre of mass already differs from the actual trajectory which would take place with the used initial condition. This is exactly the same with the orientation quaternion since the integration affects the pose and hence both, position and orientation. Eqally, the sequence values for the velocities and angular velocities have nothing to do with the intended stepped course and undergo a continuous change instead.

\subsection{Discussion}
\label{subsec:Discussion}

First of all it can be said that there is a clear training effect in all architectures, which confirms the basic function of the proposed dual quaternion neural networks. As expected, their training process and performance is decisively affected by the chosen hyper-parameter, hence the two-staged approach to determine them was really insightful and brought up the best combination of the parameters under consideration.

To comment specifically on the two different attention concepts and the corresponding experiment results: even though the binary approach theoretically has the easier task, the results can't compete with the multi-classifier. One possible explanation for this are the imbalances in the dataset, although each class is existent with a roughly equal quantity. Only $\frac{1}{7}$ of the inputs to the shared neural network correspond to the desired output one. This could be targeted with weighting the losses according to this distribution. Another problem is the fact that there is no distinct output for the case "no collision to be expected" which has to be made up with a threshold that a predicted value has to exceed to be rated as a positive prediction. Making this also a learnable parameter could potentially target this to bring the threshold to an optimal value and is worth further experiments and investigation.
The confusion matrices show that the most problematic prediction was that of "no collision to be expected". This could be because of the minor differences in the time steps just before or just after a collision with the wall occurs. To target this and to support the training process especially for these cases, the amount of input data for these cases could be increased and uncritical data instead left out as a compensation. This would mean that the strategy to keep balance in the dataset has to be changed. If this leads to higher accuracies and therefore success, this would confirm the hypothesis.

The architecture without attention suffers from major difficulties predicting the free movement through the simulation area. The capabilities regarding predicting collisions with the wall can't even be rated since no collision seems to have happened. The poor performance could be explained with the high imbalances in the dataset, even though the simulation area was chosen comparable small to make up for this as much as possible. Furthermore, the net has to perform two completely different tasks which are fundamentally different from each other: on the one hand it has to be detected that no collision will occur and therefore the twist has to be calculated and the impulse and angular momentum needs to stay unchanged, and on the other hand when a collision takes place all three values have to change after completely different principles. These may conflict with each other, resulting in the behaviour visible in the sequence prediction. 

When looking at the version with the upstream attention, the advantage of this approach becomes visible: if the case  "no collision" is correctly detected, the result is as accurate as the numerical integration allows. Furthermore, the potential disadvantage of the previous version is split into two separate tasks to simplify the overall problem. If a false prediction in the attention network occurs, this directly leads to potentially serious issues for the sequence. Either a collision with a wall is mistakenly not recognised, resulting in the rigid body leaving the simulation environment or the collision network has to predict a "false collision" where the rigid body does not hit a wall. In both cases, predictions have to be done in a domain which the network is not trained on such that no reasonable prediction can be expected. An attempt to increase the prediction capabilities for the collision network would be to change the strategy for attending to the relevant wall. Instead of leaving out the walls which are not relevant, they could be masked with a zero dual quaternion.  These zero-inputs don't bring additional information to the network, but they change the structure of the input layer and would lead to different results in the hidden layers which might be beneficial and hence is worth further investigation.

Overall, however, the performance of the architecture with attention looks promising, since in sections the trajectories are accurate and the velocities and angular velocities correctly stay constant. Optimizations in the collision detection will directly improve the prediction results.

\section{Conclusion}

In this paper we present a novel neural network architecture based on the mathematical model of dual quaternions: the dual quaternion feed forward neural networks networks. This is supplemented by the development of the dual quaternion attention mechanism. These derived general concepts are applicable on a wide range of problems which benefit from a problem description in dual quaternion space. Also the potential necessity of unit preservation is considered.

The developed dual quaternion feed forward neural networks and the dual quaternion attention were applied on the problem of predicting the dynamics of rigid bodies in a fixed simulation environment, where with the help of a hard coded recurrence and a numerical integration a sequence of rigid body poses was generated. With the usage of dual quaternions, especially suitable for describing the rigid body motions, and further physical and mathematical relations, a strong inductive bias was incorporated. For future work, improvements on the overall prediction capability, the extension to multiple rigid bodies within the simulation environment and the incorporation of external actors is planned.

\medskip

\small

\printbibliography

\appendix
\section{Appendix}
% Notationsverzeichnis
\setlength{\nomlabelwidth}{1.15cm}

\nomenclature[01]{$\times$}{Cross product}
\nomenclature[02]{$\cdot$}{Dot product}
\nomenclature[03]{$\circ$}{Hadamard product}

\nomenclature[04]{$\epsilon$}{Dual unit}

\nomenclature[05]{$\mathbb{R}$}{Real numbers}
\nomenclature[06]{$\mathbb{C}$}{Complex numbers}
\nomenclature[07]{$\mathbb{H}$}{Quaternions}
\nomenclature[08]{$\mathbb{H}_{1}$}{Unit quaternions}
\nomenclature[09]{$\mathbb{H}_{d}$}{Dual quaternions}
\nomenclature[10]{$\mathbb{H}_{d}^{1}$}{Unit dual quaternions}

\nomenclature[11]{$Q$}{Quaternion}
\nomenclature[12]{$Q^{*}$}{Quaternion conjugate}
\nomenclature[13]{$[[Q]]_{L}$}{Matrix representation of a quaternion for quaternion multiplication}
\nomenclature[14]{$[[Q]]_{R}$}{Matrix representation of a quaternion for quaternion multiplication with permuted order of sequence}

\nomenclature[15]{$Q_{d}$}{Dual quaternion}
\nomenclature[16]{$Q_{\epsilon}$}{Dual part of dual quaternion}
\nomenclature[17]{$\mathbf{q}$}{Complex vector part of a quaternion or of the real part of a dual quaternion}
\nomenclature[18]{$\mathbf{q}_{\epsilon}$}{Complex vector part of the dual part of a dual quaternion}
\nomenclature[19]{$Q_{d}^{*}$}{Dual quaternion conjugate}
\nomenclature[20]{$\bar{Q}_{d}^{*}$}{Dual quaternion dual conjugate}

\nomenclature[21]{$Q_{d}^{g}$}{Vector set of dual quaternions}
\nomenclature[22]{$\mathbf{Q}_{d}^{g}$}{Matrix set of dual quaternions}

\nomenclature[23]{$[[Q_{d}]]_{L}$}{Matrix representation of a dual quaternion for dual quaternion multiplication}
\nomenclature[24]{$[[Q_{d}]]_{R}$}{Matrix representation of a dual quaternion for dual quaternion multiplication with permuted order of sequence}

\nomenclature[25]{$Q'$}{Rotated quaternion}
\nomenclature[26]{$Q_{d}'$}{Transformed dual quaternion}

\nomenclature[27]{$\hat Q_{d}$}{Predicted dual quaternion}

\printnomenclature

\end{document}